\documentclass{article}

\usepackage{preprint,times}

\usepackage[utf8]{inputenc}
\usepackage[T1]{fontenc}
\usepackage{booktabs}
\usepackage{tabularx}
\usepackage{amsfonts}
\usepackage{amsmath}
\usepackage{microtype}
\usepackage[table]{xcolor}
\usepackage{graphicx}
\graphicspath{{Figures/}}
\usepackage{xspace}
\usepackage{float}
\usepackage{wrapfig}
\usepackage{hyperref}
\usepackage{url}

\newcommand{\method}{Premover\xspace}
\definecolor{citationborder}{RGB}{214,254,208}
\definecolor{referenceborder}{RGB}{247,206,205}
\definecolor{urlborder}{RGB}{208,254,254}
\hypersetup{
  colorlinks=false,
  linkbordercolor=referenceborder,
  citebordercolor=citationborder,
  urlbordercolor=urlborder,
  pdfborder={0 0 0.5}
}
\DeclareRobustCommand{\linkedref}[2]{#1~\ref{#2}}
\title{Premover: Fast Vision-Language-Action Control via Early Execution During Instruction Delivery}

\author{
  Joonha Park \quad Jiseung Jeong \quad Taesik Gong\thanks{Corresponding author.} \\
  UNIST, Ulsan, Republic of Korea \\
  \texttt{\{joonhapark,wjdwltmd1151,taesik.gong\}@unist.ac.kr}
}

\begin{document}

\setcounter{footnote}{1}
\maketitle

\begin{abstract}
Vision-Language-Action (VLA) policies are typically evaluated under the assumption that the robot starts acting only after the user has finished typing or speaking. In real interactions, however, entering an instruction can take several seconds, leaving the policy idle for a substantial fraction of the interaction. Partial instructions may already contain sufficient information to begin acting before the full instruction arrives. We introduce \textbf{Premover}, a parameter-lightweight module that reduces interaction latency by overlapping instruction delivery with robot execution while keeping the VLA backbone frozen. Premover uses a learned \emph{vision-language focus map} to identify \emph{where} the current partial instruction refers in the visual scene, and an \emph{action readiness gate} to determine \emph{when} execution can begin. On the SO-101 robot with speech and online transcription, Premover reduces mean wall-clock time by $17.9\%$, from $60.5$ s to $49.7$ s, while maintaining task success rates. In simulation on LIBERO with $\pi_{0.5}$ at the speaking rate, Premover reduces mean wall-clock time by $8.3\%$ while maintaining comparable success to waiting for the complete instruction before execution.
\end{abstract}

\section{Introduction}
\label{sec:introduction}

\begin{wrapfigure}{r}{0.38\linewidth}
    \vspace{-16pt}
    \centering
    \includegraphics[width=\linewidth]{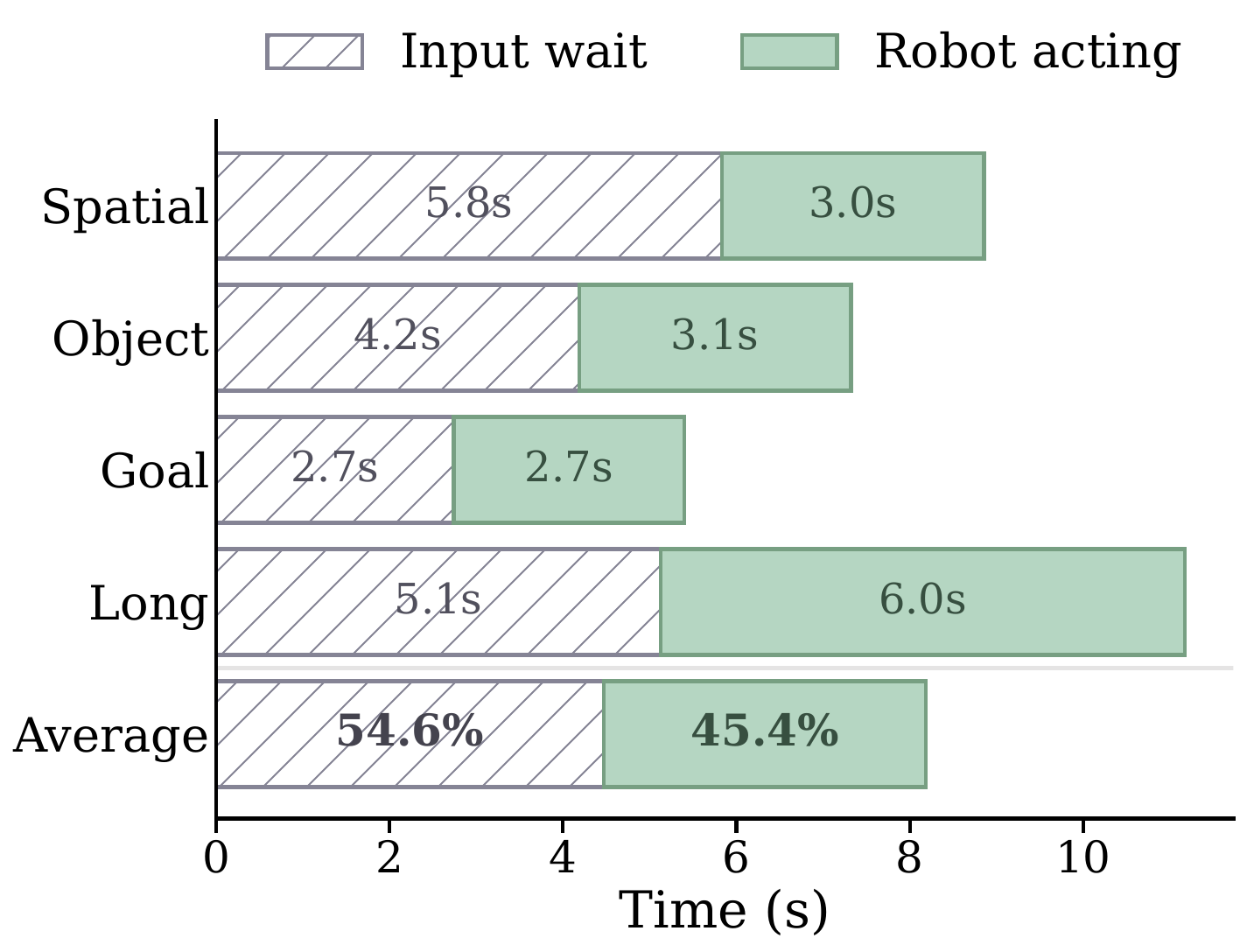}
    \caption{\textbf{Breakdown of waiting and action time on LIBERO.} Hatched bars show measured instruction delivery time; solid bars show measured robot action time at the speaking rate.}
    \label{fig:instructionT}
    \vspace{-10pt}
\end{wrapfigure}

Recent work on efficient Vision-Language-Action (VLA) models focuses on reducing inference latency once the complete instruction is available~\citep{yue2024deervla,liu2025vlapruner,wen2024tinyvla,pertsch2025fast}. In real-world deployment, however, the instruction takes time to arrive. A user may take several seconds to type or speak a request, during which the policy typically remains idle. This instruction-delivery interval accounts for approximately $54.6\%$ of the wall-clock time across the four LIBERO suites at the speaking rate (\linkedref{Figure}{fig:instructionT}). Yet standard evaluation protocols~\citep{liu2023libero, zhang2025vlaarena} largely ignore this delay by presenting the complete instruction before execution begins.

We ask whether some of this instruction-delivery interval can instead be used for robot execution. In our setting, instruction words arrive incrementally, and we refer to the portion received so far as the \emph{streaming prefix}. Even a streaming prefix such as ``Pick the tomato on top of the drawer\dots'' may already contain sufficient information to begin execution, allowing instruction delivery and robot execution to overlap. The challenge is that beginning execution too early can lead the robot to take actions that are not yet supported by the current streaming prefix. This raises two questions. First, \emph{where} does the current streaming prefix refer in the visual scene? Second, \emph{when} does the streaming prefix contain sufficient information to begin execution?

We introduce \textbf{\method}, a lightweight module built on a frozen VLA policy. \linkedref{Figure}{fig:teaser} contrasts \method{} with conventional inference, where execution begins only after the complete instruction is received. \method{} addresses these questions through two components: a \emph{vision-language focus map} that identifies \emph{where} the current streaming prefix refers in the visual scene, and an \emph{action readiness gate} that determines \emph{when} the prefix contains sufficient information to begin execution. Both components are trained jointly on features from the frozen VLA backbone. After execution begins, the instruction continues to arrive, and \method{} updates its action prediction as the streaming prefix grows.

\begin{figure}[t]
    \centering
    \includegraphics[width=\linewidth]{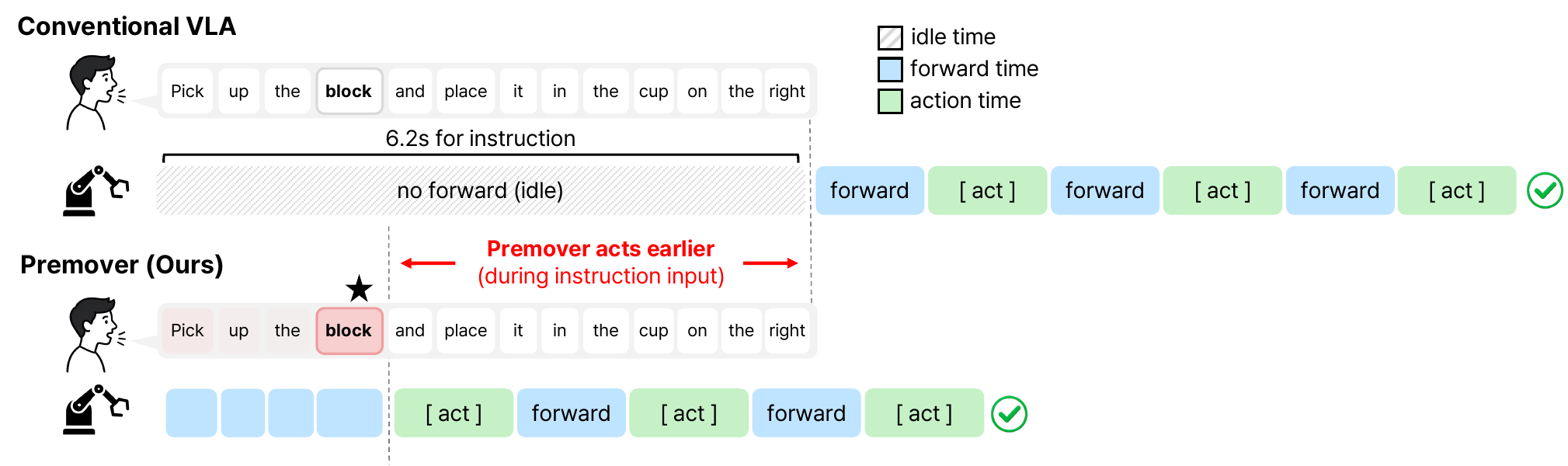}
    \caption{\textbf{Conventional VLA inference vs.\ \method.} Conventional VLAs wait for the complete instruction before execution. \method{} processes the streaming prefix as it arrives, using a vision-language focus map to determine where to focus and an action readiness gate to determine when to begin.}
    \label{fig:teaser}
\end{figure}

We evaluate \method{} with two frozen VLA backbones in simulation, $\pi_{0.5}$~\citep{black2025pi05} and GR00T-N1.7~\citep{nvidia2026gr00tn17}, on LIBERO~\citep{liu2023libero} and VLA-Arena~\citep{zhang2025vlaarena}. These benchmarks cover different task distributions, scene compositions, and instruction complexity. We evaluate both backbones under two instruction-delivery rates, a typing rate and a speaking rate, using representative rates estimated through the empirical analysis in \linkedref{Appendix}{app:input_rate}. We further deploy \method{} with $\pi_{0.5}$ on an SO-101 robot using a TTS-ASR streaming pipeline at the speaking rate.

On the SO-101 robot at the speaking rate, \method{} reduces mean wall-clock time by $17.9\%$ relative to waiting for the complete instruction before execution, with mean success rates of $86.0\%$ and $81.0\%$, respectively. On LIBERO with $\pi_{0.5}$ at the speaking rate, \method{} reduces mean wall-clock time by $8.3\%$, from $8.17$ to $7.49$ seconds, while maintaining comparable success ($97.6\%$ versus $97.9\%$). Together, these results show that instruction-delivery time, a previously overlooked source of end-to-end latency, can be reduced by overlapping it with robot execution while preserving task performance.

\section{Related Work}
\label{sec:related}

Vision-Language-Action (VLA) models adapt pretrained vision-language models to robotic control. Early approaches predict robot actions as autoregressive action-token prediction~\citep{brohan2023rt2,kim2024openvla}, while more recent architectures combine vision-language backbones with diffusion- or flow-matching-based action experts to generate continuous action chunks~\citep{chi2023diffusion,zhao2023act,black2024pi0,black2025pi05,nvidia2025groot,shukor2025smolvla}. Other work focuses on preserving or enriching the visual representations available to VLA policies. Knowledge Insulation~\citep{driess2025knowledge} preserves pretrained vision-language representations by blocking gradients from the action expert, while RoboGround~\citep{huang2025roboground} and VAP~\citep{lee2025vap} provide explicit spatial guidance through grounded masks or object-specific visual cues.
    
Most VLA methods assume that the complete instruction is available before execution begins. Earlier work on incremental language understanding in human-robot interaction instead allows a robot to process and act on an utterance before the speaker has finished~\citep{schlangen2009general,brick2007incremental}. These systems, however, rely on task-specific semantic parsers or hand-designed controllers rather than pretrained, frozen VLA backbones. \method{} builds on this incremental setting while operating directly with frozen VLA policies.

A separate line of work reduces VLA inference and execution latency through early exiting~\citep{yue2024deervla}, adaptive visual-token pruning~\citep{pei2026action}, compact multimodal backbones~\citep{wen2024tinyvla}, more efficient action tokenization~\citep{pertsch2025fast}, or adaptive action-chunk execution~\citep{wang2026vla}. These approaches improve the efficiency of producing or executing each policy decision, but still primarily operate after the complete instruction is available. In contrast, \method{} targets the instruction-delivery interval itself by beginning execution before the full instruction arrives, and can therefore be combined with methods that improve VLA inference or execution efficiency.


\section{Method}
\label{sec:methods}

\begin{figure}[t]
    \centering
    \includegraphics[width=\linewidth]{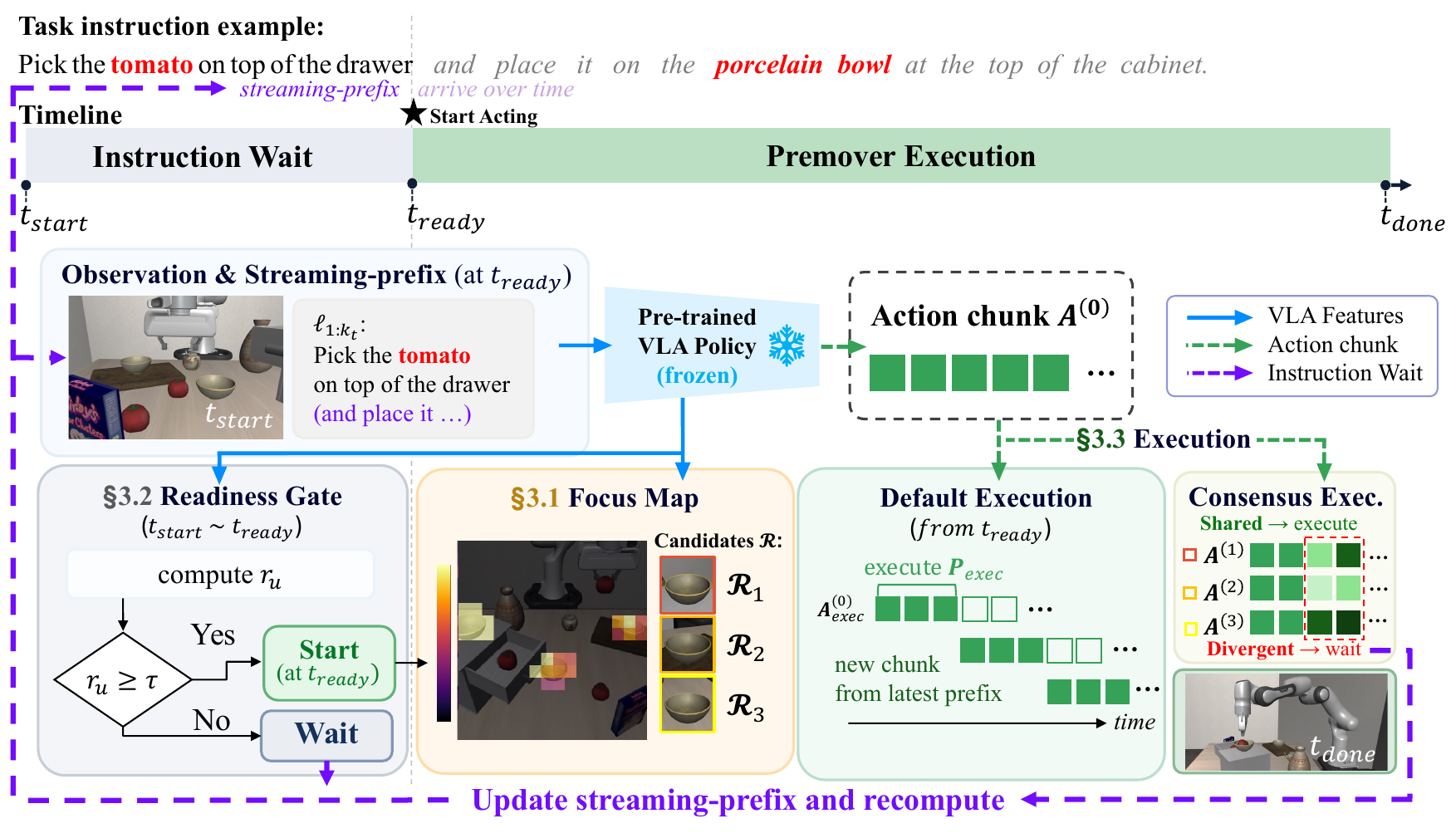}
    \caption{\textbf{Overview of \method.} The vision-language focus map (\hyperref[sec:focus]{\S\ref*{sec:focus}}) identifies image regions relevant to the current streaming prefix, while the action readiness gate (\hyperref[sec:method_readiness]{\S\ref*{sec:method_readiness}}) determines when the prefix contains sufficient information to begin execution, defining $t_{\mathrm{ready}}$. As additional words arrive, the policy updates its action prediction using the growing streaming prefix and continues execution as more information becomes available.}
    \label{fig:premover}
\end{figure}

\paragraph{Motivation and challenges.} Conventional VLA inference assumes that the complete instruction is available before the robot begins execution. In real-world interaction, however, the robot remains idle for several seconds while the user enters an instruction. On LIBERO-Spatial at speaking rate, waiting for the complete instruction takes an average of $5.8$ seconds, accounting for approximately $65.9\%$ of the wall-clock time. Standard VLA evaluation protocols ignore this idle time. This raises a natural question: \emph{can the policy begin execution before the user has finished entering the instruction?} We refer to the partial instruction observed so far as the \emph{streaming prefix}.

Leveraging the streaming prefix requires resolving two challenges. \emph{(i) Where to focus in the visual scene:} a streaming prefix such as ``Pick the tomato on top of the drawer\dots'' may already contain sufficient information to begin execution before the complete instruction arrives. However, a frozen VLA backbone does not directly localize the referent of the current streaming prefix, and its vision-language alignment can remain spread across irrelevant image regions (\linkedref{Appendix}{app:focus-map}). \emph{(ii) When to start acting:} acting on a streaming prefix can cause the robot to grasp the wrong object or place it in the wrong location, leading to task failure. The policy must therefore determine when the current streaming prefix contains sufficient information to begin execution.

\paragraph{Method overview.} \linkedref{Figure}{fig:premover} illustrates \method{}, a lightweight module that addresses these two challenges while keeping the VLA backbone frozen. The \emph{vision-language focus map} identifies \emph{where} the current streaming prefix refers in the visual scene, while the \emph{action readiness gate} determines \emph{when} the streaming prefix contains sufficient information to begin execution. Once the action readiness gate determines that execution can begin, the robot executes the predicted action chunk while the instruction continues to arrive. If subsequent actions still depend on information that has not yet arrived, \method{} can temporarily restrict execution by comparing action predictions across candidate regions.

\paragraph{Problem formulation.} 
We consider a VLA setting in which the instruction arrives incrementally rather than being provided in full at the start. At step $t$, let $\ell_{1:k_t}$ denote the first $k_t$ instruction words received so far. Tokenizing this streaming prefix yields $L_t$ language tokens. Let $N$ denote the total number of image patches across all input camera views. Given the streaming prefix, current image observations, and proprioceptive state, the policy predicts a default action chunk $A^{(0)}=(a_1^{(0)},\ldots,a_M^{(0)})$ for the next $M$ steps. The robot executes only the first $P_{\mathrm{exec}} \leq M$ actions from each predicted action chunk. The key is to determine both where the streaming prefix refers in the visual scene and when it contains sufficient information to begin execution.

\subsection{Vision-Language Focus Map}
\label{sec:focus}

Beginning execution from a streaming prefix requires identifying which image regions the current streaming prefix refers to. We therefore introduce a supervised vision-language focus map that assigns a relevance score to each image patch based on the current streaming prefix.

\paragraph{Focus Score.}
Let $h_{\mathrm{img}}^b$ and $h_{\mathrm{lang}}^b$ denote the hidden states of image patches and language tokens extracted from backbone layer $b=1,\ldots,B$.
We use trainable projection heads $f_{\mathrm{img}}$ and $f_{\mathrm{lang}}$ to map the image and language features into a shared latent space.
We assign a trainable layer-mixing logit $\gamma_b$ to each backbone layer and obtain $\omega_b=\operatorname{softmax}(\gamma)_b$.
The same layer weights $\omega_b$ are shared across the image and language modalities:
\begin{equation}
\begin{aligned}
z_{\mathrm{img}}&=\operatorname{norm}\left(\sum_{b=1}^{B}\omega_b f_{\mathrm{img}}(h_{\mathrm{img}}^b)\right),\\
z_{\mathrm{lang}}&=\operatorname{norm}\left(\sum_{b=1}^{B}\omega_b f_{\mathrm{lang}}(h_{\mathrm{lang}}^b)\right).
\end{aligned}
\label{eq:focus-features}
\end{equation}
The projection functions and $\ell_2$ normalization are applied independently to each image patch and language token.

For each image patch $i$, we compare its projected feature $z_{\mathrm{img},i}$ with each token feature $z_{\mathrm{lang},j}$, where $j$ indexes the language tokens in the streaming prefix, and retain the strongest match:

\begin{equation}
p_i=\sigma\left(\beta\max_{1\leq j\leq L_t}\left\langle z_{\mathrm{img},i},z_{\mathrm{lang},j}\right\rangle\right),
\label{eq:focus-prob}
\end{equation}
where $\sigma$ is the sigmoid function and $\beta$ is a fixed similarity scale. The resulting $p=(p_1,\ldots,p_N)$ is the vision-language focus map. Each $p_i$, which we call the \emph{focus score}, measures the strongest vision-language match between image patch $i$ and the tokens in the current streaming prefix.

To condition the action prediction on a region selected by the focus map, we convert each selected region $R$ into soft visual weights,
$w_i=\alpha+(1-\alpha)\mathbf{1}[i\in R]$. Patches inside $R$ receive full weight, while patches outside are suppressed to weight $\alpha\in[0,1)$ rather than discarded. The policy then predicts the action chunk using these reweighted visual features.

The focus map is supervised using two objectives. The focus-map loss ($\mathcal{L}_{\mathrm{focus}}$) matches individual patch focus scores to segmentation-mask labels. The region-contrastive loss ($\mathcal{L}_{\mathrm{contrast}}$) assigns higher mean focus scores to prefix-relevant regions than to other visible regions.     Further details are provided in \linkedref{Appendix}{app:impl}.

\subsection{Action Readiness Gate}
\label{sec:method_readiness}

The action readiness gate determines whether the current streaming prefix contains sufficient information to begin execution. Let $e_u \in \{0,1\}$ denote the binary readiness decision at streaming prefix update $u$. Here, $e_u=1$ means that the current streaming prefix is sufficiently informative to support an initial action, whereas $e_u=0$ indicates that more information is needed before execution can begin.

\paragraph{Readiness score.}
Let $x_u$ denote a representation of the current observation and streaming prefix, built from backbone hidden states. A two-layer readiness predictor $f_{\mathrm{ready}}$ maps $x_u$ to a scalar logit, from which we compute:
\begin{equation}
r_u=\sigma(f_{\mathrm{ready}}(x_u)),\qquad
e_u=\mathbf{1}[r_u\geq\tau],
\label{eq:readiness-score}
\end{equation}
where $\sigma$ is the sigmoid function and $\tau$ is a learned threshold. Execution begins when $e_u{=}1$ at two consecutive streaming-prefix updates with distinct word counts. We jointly train $f_{\mathrm{ready}}$ and $\tau$ using the readiness loss, $\mathcal{L}_{\mathrm{ready}}$, a weighted binary cross-entropy objective with prefix-level labels derived from a \emph{readiness cutoff} annotated by a language model, defined as the earliest word sufficient to identify the target and support the initial action. Backbone-specific feature aggregation, label construction, and the detailed training objective are provided in \linkedref{Appendix}{app:impl}.

\subsection{Execution}
\label{sec:method_execution}

Once the action readiness gate determines that the current streaming prefix contains sufficient information to begin execution, the robot begins acting from the default action chunk $A^{(0)}$ before the complete instruction arrives. As additional words arrive, the policy updates its action prediction using the current observation and updated streaming prefix.

\paragraph{Default action execution.}
\label{sec:default_execution}

By default, the robot executes a fixed number of actions from each predicted action chunk. Given the default action chunk $A^{(0)}=(a_1^{(0)},\ldots,a_M^{(0)})$, we define the default execution length as
\begin{equation}
m_{\mathrm{exec}}=\min(P_{\mathrm{exec}},M)=P_{\mathrm{exec}},
\label{eq:default_execution}
\end{equation}
where the final equality follows from $P_{\mathrm{exec}}\leq M$. The robot then executes the first $m_{\mathrm{exec}}$ actions,
\begin{equation}
A_{\mathrm{exec}}^{(0)}
=
A_{1:m_{\mathrm{exec}}}^{(0)}
=
\left(a_1^{(0)},\ldots,a_{m_{\mathrm{exec}}}^{(0)}\right).
\end{equation}
After executing these actions, the policy predicts a new action chunk using the current observation and the latest streaming prefix. Under default execution, each prediction therefore provides up to $P_{\mathrm{exec}}$ actions before replanning. If the current streaming prefix does not yet provide enough information for later actions, \method{} can execute fewer than $P_{\mathrm{exec}}$ actions and wait for additional instruction. We handle this case with consensus-guided execution, described below.

\paragraph{Consensus-guided execution.}
\label{sec:method_consensus}

Consensus-guided execution is used when later actions still depend on information that has not yet arrived. For example, ``pick the tomato on top of the drawer and \dots'' identifies the object but not its destination (\linkedref{Figure}{fig:hedge_commit}). \method{} extracts $J$ candidate regions from the focus map, obtains a region-conditioned action chunk $A^{(d)}$ for each using the reweighting in \linkedref{Section}{sec:focus}, and executes only the initial actions shared across these chunks. Let $S(\mathcal{A})$ denote the length of this shared action prefix, and let $m_{\mathrm{exec}}^{\mathrm{cons}}$ denote the number of actions executed under consensus-guided execution:
\begin{equation}
m_{\mathrm{exec}}^{\mathrm{cons}}
=
\min\left(
S\left({A^{(1)},\ldots,A^{(J)}}\right),
C,
P_{\mathrm{exec}}
\right),
\label{eq:premove_burst}
\end{equation}

where $C$ limits the number of actions executed under consensus-guided execution. Further details are provided in \linkedref{Appendix}{app:consensus_details}.

\subsection{Training Objective}
\label{sec:premove_training}
\label{sec:method_objective}

We jointly train the image and language projection heads, shared layer-mixing logits $\gamma$, and the action readiness gate using
\begin{equation}
\mathcal{L}=\mathcal{L}_{\mathrm{focus}}+\mathcal{L}_{\mathrm{contrast}}+\mathcal{L}_{\mathrm{ready}},
\label{eq:total-loss}
\end{equation}
where $\mathcal{L}_{\mathrm{focus}}$ provides patch-level focus supervision, $\mathcal{L}_{\mathrm{contrast}}$ separates prefix-relevant regions from other visible regions, and $\mathcal{L}_{\mathrm{ready}}$ supervises the readiness decision. The vision-language focus map and action readiness gate are trained jointly, while the VLA backbone remains frozen. Training settings, loss details, data splits, and optimization hyperparameters are provided in \linkedref{Appendix}{app:impl}.

\section{Experiments}
\label{sec:experiments}

\subsection{Experimental Setup}
\label{sec:exp_setup}

\begin{wrapfigure}{r}{0.4\linewidth}
    \vspace{-15pt}
    \centering
    \includegraphics[width=\linewidth]{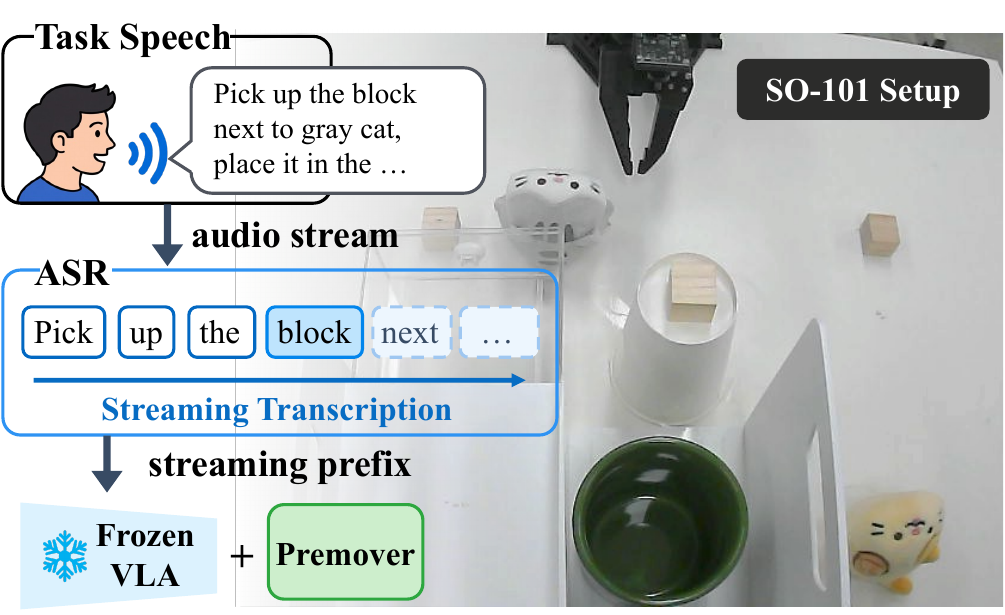}
    \caption{\textbf{Real-robot system pipeline.}
    An ASR model streams partial transcripts, which \method{} combines with robot observations to produce action chunks before the full instruction arrives.}
    \label{fig:real_pipeline}
    \vspace{-15pt}
\end{wrapfigure}

\paragraph{Evaluation settings.}
We evaluate \method{} in both simulation and on a real robot with streaming instructions. Under matched configurations, we compare \emph{Premover} with \emph{Full-prompt}, which begins execution after the complete instruction arrives; \emph{Right-away}, which begins execution immediately; and \emph{Noun-triggered}, which begins execution when the target noun arrives. For the speaking condition, both simulation and real-robot experiments use TTS-generated speech with ASR transcripts. We report success rate (SR) for simulation and real-robot single-object tasks, partial success scores for real-robot disambiguation tasks, and mean wall-clock time over all episodes (Wall). Protocol and timing details are provided in \linkedref{Appendix}{app:input_rate}.

\paragraph{Real-robot environments.}
\begin{table*}[!t]
\centering
\scriptsize
\caption{
\textbf{Real-robot evaluation with spoken instructions}. On five single-object tasks, we report full-task success rate (SR; \%), while on three disambiguation tasks, we report partial success score (\%). We also report mean wall-clock time (Wall; s), with $20$ trials per task and method and a $900$-step limit. Avg. reports the mean across tasks within each group.
}
\label{tab:real_robot}
\setlength{\tabcolsep}{2.5pt}

\resizebox{\linewidth}{!}{%
\begin{tabular}{lcccccc|cccc}
\toprule
& \multicolumn{6}{c|}{\textbf{Single-object}}
& \multicolumn{4}{c}{\textbf{Disambiguation}} \\
\cmidrule(lr){2-7}
\cmidrule(lr){8-11}

\textbf{Method}
& \textbf{Block$\rightarrow$Cup}
& \textbf{Drawer}
& \textbf{Cat$\rightarrow$Cup}
& \textbf{Cup$\rightarrow$Cup}
& \textbf{Push Cup}
& \textbf{Avg.}
& \textbf{On Cup}
& \textbf{Gray Cat}
& \textbf{Orange Cat}
& \textbf{Avg.} \\
\midrule

\multicolumn{11}{l}{\textbf{(a) Success rate (\%) $\uparrow$}} \\

Full-prompt
& 65.0 & 70.0 & 95.0 & 90.0 & 85.0 & 81.0
& 52.5 & 55.0 & 65.0 & 57.5 \\

Right-away
& 50.0 & 55.0 & 85.0 & 70.0 & 70.0 & 66.0
& 25.0 & 45.0 & 35.0 & 35.0 \\

Noun-triggered
& 40.0 & 55.0 & 65.0 & 40.0 & 75.0 & 55.0
& 32.5 & 30.0 & 40.0 & 34.2 \\

\rowcolor{green!10}
\textbf{\method{} (Ours)}
& 75.0 & 80.0 & 90.0 & 95.0 & 90.0 & 86.0
& 57.5 & 90.0 & 67.5 & 71.7 \\

\midrule
\multicolumn{11}{l}{\textbf{(b) Wall-clock time (s) $\downarrow$}} \\

Full-prompt
& 66.6 & 79.1 & 48.4 & 53.5 & 54.9 & 60.5
& 83.6 & 79.3 & 101.8 & 88.2 \\

Right-away
& 60.7 & 101.4 & 48.6 & 54.0 & 63.0 & 65.5
& 83.8 & 80.0 & 101.3 & 88.4 \\

Noun-triggered
& 75.2 & 97.2 & 57.7 & 88.4 & 65.7 & 76.8
& 96.8 & 137.2 & 99.0 & 111.0 \\

\rowcolor{green!10}
\textbf{\method{} (Ours)}
& 49.0 & 73.2 & 38.8 & 32.9 & 54.7 & 49.7
& 77.8 & 71.0 & 81.2 & 76.7 \\

\bottomrule
\end{tabular}
}
\end{table*}

We evaluate the frozen $\pi_{0.5}$ backbone~\citep{black2025pi05} on an SO-101 robot (\linkedref{Figure}{fig:real_pipeline}). The evaluation comprises five single-object tasks and three disambiguation variants of a block-to-drawer task. Single-object tasks require moving one object to a destination, while disambiguation tasks require selecting the instructed object from multiple identical objects and moving it to a destination. A frozen Segment Anything Model (SAM)~\citep{kirillov2023sam} provides offline mask supervision for the vision-language focus map. Details are provided in \linkedref{Appendix}{app:real_robot}.

\begin{table*}[!t]
\centering
\scriptsize
\setlength{\tabcolsep}{1.8pt}
\renewcommand{\arraystretch}{1.05}
\caption{\textbf{Main results on LIBERO with $\pi_{0.5}$.}
Success rate (SR; \%) and wall-clock time (Wall; s) across the four LIBERO suites at both instruction-delivery rates. The two rates correspond to representative typing and speaking speeds (\linkedref{Appendix}{app:input_rate}). Entries are means over three evaluation seeds, with seed-level standard deviations reported for both SR and Wall.}
\label{tab:libero_pi05}

\begin{tabularx}{\textwidth}{@{}l *{10}{>{\centering\arraybackslash}X}@{}}
\toprule
& \multicolumn{2}{c}{\textbf{Spatial}}
& \multicolumn{2}{c}{\textbf{Object}}
& \multicolumn{2}{c}{\textbf{Goal}}
& \multicolumn{2}{c}{\textbf{Long-10}}
& \multicolumn{2}{c}{\textbf{Mean}} \\
\cmidrule(lr){2-3}
\cmidrule(lr){4-5}
\cmidrule(lr){6-7}
\cmidrule(lr){8-9}
\cmidrule(lr){10-11}

\textbf{Method}
& SR $\uparrow$ & Wall $\downarrow$
& SR $\uparrow$ & Wall $\downarrow$
& SR $\uparrow$ & Wall $\downarrow$
& SR $\uparrow$ & Wall $\downarrow$
& SR $\uparrow$ & Wall $\downarrow$ \\
\midrule

\multicolumn{11}{l}{\textit{Typing rate}} \\
Full-prompt
& 99.1{\tiny$\pm$0.3} & 20.27{\tiny$\pm$0.02}
& 99.3{\tiny$\pm$0.3} & 14.70{\tiny$\pm$0.02}
& 97.7{\tiny$\pm$0.3} & 10.14{\tiny$\pm$0.01}
& 94.3{\tiny$\pm$0.9} & 20.59{\tiny$\pm$0.05}
& 97.6{\tiny$\pm$0.4} & 16.42{\tiny$\pm$0.02} \\

Right-away
& 70.9{\tiny$\pm$1.5} & 9.46{\tiny$\pm$0.35}
& 73.2{\tiny$\pm$1.3} & 7.60{\tiny$\pm$0.23}
& 56.4{\tiny$\pm$2.2} & 10.08{\tiny$\pm$0.21}
& 85.2{\tiny$\pm$1.3} & 9.34{\tiny$\pm$0.23}
& 71.4{\tiny$\pm$0.7} & 9.12{\tiny$\pm$0.12} \\

Noun-triggered
& 86.0{\tiny$\pm$1.4} & 10.18{\tiny$\pm$0.22}
& 99.6{\tiny$\pm$0.2} & 7.56{\tiny$\pm$0.02}
& 79.9{\tiny$\pm$1.0} & 8.34{\tiny$\pm$0.08}
& 95.0{\tiny$\pm$1.2} & 10.87{\tiny$\pm$0.14}
& 90.1{\tiny$\pm$0.3} & 9.24{\tiny$\pm$0.03} \\

\rowcolor{green!10}
\textbf{\method{}}
& 97.6{\tiny$\pm$1.0} & 16.44{\tiny$\pm$0.16}
& 98.8{\tiny$\pm$0.4} & 11.83{\tiny$\pm$0.13}
& 98.0{\tiny$\pm$0.9} & 9.88{\tiny$\pm$0.06}
& 95.1{\tiny$\pm$1.9} & 16.61{\tiny$\pm$0.17}
& 97.4{\tiny$\pm$0.5} & 13.69{\tiny$\pm$0.02} \\

\midrule

\multicolumn{11}{l}{\textit{Speaking rate}} \\
Full-prompt
& 98.7{\tiny$\pm$0.3} & 8.84{\tiny$\pm$0.01}
& 99.6{\tiny$\pm$0.4} & 7.31{\tiny$\pm$0.02}
& 97.9{\tiny$\pm$1.3} & 5.45{\tiny$\pm$0.06}
& 95.4{\tiny$\pm$1.0} & 11.07{\tiny$\pm$0.09}
& 97.9{\tiny$\pm$0.1} & 8.17{\tiny$\pm$0.02} \\

Right-away
& 77.0{\tiny$\pm$1.2} & 6.10{\tiny$\pm$0.07}
& 65.0{\tiny$\pm$1.1} & 6.62{\tiny$\pm$0.05}
& 64.6{\tiny$\pm$1.8} & 6.36{\tiny$\pm$0.09}
& 86.0{\tiny$\pm$0.5} & 8.39{\tiny$\pm$0.03}
& 73.1{\tiny$\pm$0.4} & 6.87{\tiny$\pm$0.02} \\

Noun-triggered
& 84.4{\tiny$\pm$0.7} & 6.89{\tiny$\pm$0.06}
& 99.6{\tiny$\pm$0.3} & 5.64{\tiny$\pm$0.02}
& 90.3{\tiny$\pm$1.0} & 4.96{\tiny$\pm$0.08}
& 95.7{\tiny$\pm$0.3} & 8.58{\tiny$\pm$0.04}
& 92.5{\tiny$\pm$0.1} & 6.52{\tiny$\pm$0.02} \\

\rowcolor{green!10}
\textbf{\method{}}
& 98.2{\tiny$\pm$0.6} & 7.78{\tiny$\pm$0.04}
& 99.2{\tiny$\pm$0.2} & 6.54{\tiny$\pm$0.02}
& 97.6{\tiny$\pm$1.2} & 5.45{\tiny$\pm$0.07}
& 95.2{\tiny$\pm$2.1} & 10.19{\tiny$\pm$0.08}
& 97.6{\tiny$\pm$0.5} & 7.49{\tiny$\pm$0.02} \\

\bottomrule
\end{tabularx}
\end{table*}

\begin{table*}[!t]
\centering
\scriptsize
\setlength{\tabcolsep}{4.5pt}
\caption{\textbf{Evaluation across backbone-benchmark combinations at the speaking rate.}
Success rate (SR; \%) and wall-clock time (Wall; s) for the remaining backbone-benchmark combinations at speaking rate.
\linkedref{Table}{tab:libero_pi05} covers $\pi_{0.5}$ on LIBERO.
Entries are means over three evaluation seeds, with seed-level standard deviations reported for both SR and Wall.
Complete results across both instruction-delivery rates and all benchmark suites are provided in \linkedref{Appendix}{app:full_results}.}
\label{tab:generalization_150}

\begin{tabularx}{\textwidth}{@{}l *{6}{>{\centering\arraybackslash}X}@{}}
\toprule
& \multicolumn{2}{c}{GR00T-N1.7 / LIBERO}
& \multicolumn{2}{c}{$\pi_{0.5}$ / VLA-Arena}
& \multicolumn{2}{c}{GR00T-N1.7 / VLA-Arena} \\
\cmidrule(lr){2-3}
\cmidrule(lr){4-5}
\cmidrule(lr){6-7}

\textbf{Method}
& \textbf{SR $\uparrow$} & \textbf{Wall $\downarrow$}
& \textbf{SR $\uparrow$} & \textbf{Wall $\downarrow$}
& \textbf{SR $\uparrow$} & \textbf{Wall $\downarrow$} \\
\midrule

Full-prompt
& 94.0{\tiny$\pm$0.3} & 8.60{\tiny$\pm$0.03}
& 64.3{\tiny$\pm$0.8} & 10.37{\tiny$\pm$0.05}
& 48.6{\tiny$\pm$1.1} & 8.12{\tiny$\pm$0.05} \\

Right-away
& 57.7{\tiny$\pm$0.9} & 8.58{\tiny$\pm$0.05}
& 60.7{\tiny$\pm$0.7} & 8.93{\tiny$\pm$0.06}
& 40.0{\tiny$\pm$0.3} & 7.27{\tiny$\pm$0.01} \\

Noun-triggered
& 77.2{\tiny$\pm$0.8} & 7.95{\tiny$\pm$0.06}
& 63.5{\tiny$\pm$0.7} & 8.97{\tiny$\pm$0.04}
& 44.4{\tiny$\pm$0.8} & 7.18{\tiny$\pm$0.05} \\

\rowcolor{green!10}
\textbf{\method{}}
& 93.8{\tiny$\pm$0.5} & 8.07{\tiny$\pm$0.05}
& 63.7{\tiny$\pm$1.3} & 9.63{\tiny$\pm$0.08}
& 48.3{\tiny$\pm$0.7} & 7.64{\tiny$\pm$0.03} \\

\bottomrule
\end{tabularx}
\end{table*}

\subsection{Real-Robot Results}
\label{sec:real_results}

\linkedref{Table}{tab:real_robot} shows real-robot results at the speaking rate. Across the five single-object tasks, \method{} reduces mean wall-clock time by $17.9\%$, from $60.5$ to $49.7$~s; the corresponding mean success rates are $86.0\%$ for \method{} and $81.0\%$ for Full-prompt. Disambiguation tasks require distinguishing the target from multiple identical objects using spatial descriptions that arrive after the object noun. On these tasks, \method{} reduces mean wall-clock time from $88.2$ to $76.7$~s while maintaining task performance, with a mean success score of $71.7\%$ compared with $57.5\%$. \method{} achieves lower wall-clock times and higher success scores than all baselines on every disambiguation task.

The real-robot results further highlight the role of the action readiness gate. With online transcription and noisier visual observations, delaying execution until the streaming prefix contains sufficient information can reduce the risk of acting on incomplete or misrecognized instructions, helping avoid the large success drop observed with Right-away. After execution begins, the focus map can influence subsequent action predictions and execution trajectories, which may also contribute to the higher success rates.

\subsection{Simulation Results}
\label{sec:exp_main}
\label{sec:exp_libero}
\label{sec:exp_cross_model}

Across all four backbone-benchmark combinations at the speaking rate, \method{} reduces mean wall-clock time by $5.9$-$8.3\%$ relative to Full-prompt while keeping success within $0.6$ percentage points of the corresponding baseline (\linkedref{Tables}{tab:libero_pi05} and~\ref{tab:generalization_150}). The consistent reduction across two VLA backbones and two benchmarks shows that the latency benefit of streaming execution extends beyond a single model or task distribution. Although Right-away and Noun-triggered can further reduce wall-clock time in some settings, they often incur substantial drops in success, highlighting the importance of determining when execution can safely begin. Full results for both instruction-delivery rates are provided in \linkedref{Appendix}{app:full_results}.

\subsection{Analysis}
\label{sec:analysis}
\begin{figure*}[t]
    \centering
    \includegraphics[width=\textwidth]{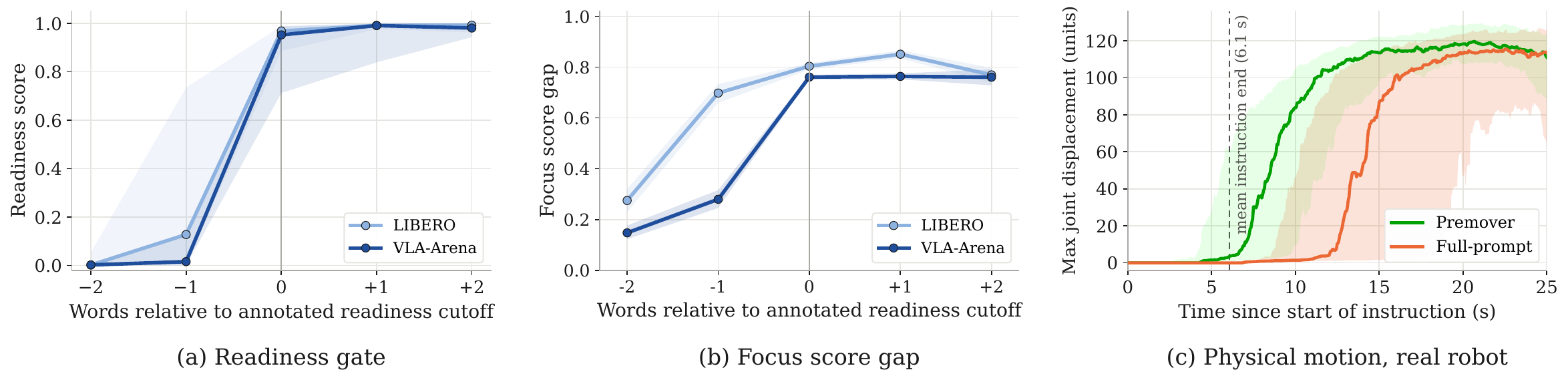}
    \caption{\textbf{Analysis of early execution.}
(a) Readiness scores, averaged over $\pi_{0.5}$ and GR00T-N1.7, increase around the annotated readiness cutoff on LIBERO and VLA-Arena.
(b) Mean focus-score gap between the top-16 and remaining patches, averaged over both backbones. The gap widens around the cutoff. (c) Averaged across all real-world tasks, \method{} begins measurable joint motion before Full-prompt, including before instruction delivery finishes.}
    \label{fig:analysis}
\end{figure*}
We analyze how \method{}'s action readiness gate and vision-language focus map behave around the annotated readiness cutoff. \linkedref{Figure}{fig:analysis}(a) shows readiness scores on LIBERO and VLA-Arena, averaged over $\pi_{0.5}$ and GR00T-N1.7. The scores remain low before the cutoff and rise sharply around it, indicating that the gate responds to when the streaming prefix becomes sufficient to support the initial action rather than to a fixed amount of observed language. \linkedref{Figure}{fig:analysis}(b) examines the vision-language focus map using the gap between the mean focus scores of the top-$K$ patches ($K=16$) and the remaining patches. The gap increases sharply around the readiness cutoff on both benchmarks, showing that the focus map becomes more spatially concentrated as the instruction provides sufficient information to act.

We then examine whether early execution translates into physical behavior on the real robot. \linkedref{Fig.}{fig:analysis}(c) shows the maximum joint displacement from the starting configuration, averaged across all real-world tasks. \method{} begins moving substantially earlier than Full-prompt, including before instruction delivery is complete, showing that early execution translates into earlier physical motion. The same trend holds for $K\in\{8,16,32\}$, with the sensitivity analysis and the definition of the joint-displacement metric provided in \linkedref{Appendix}{app:focus_k_sensitivity}.

\subsection{Ablation Studies}
\label{sec:exp_ablation}
\subsubsection{Effect of Instruction-Delivery Rate}
\label{sec:abl_rate}

\begin{wrapfigure}{r}{0.48\textwidth}
    \vspace{-1.3em}
    \centering
    \includegraphics[width=\linewidth]{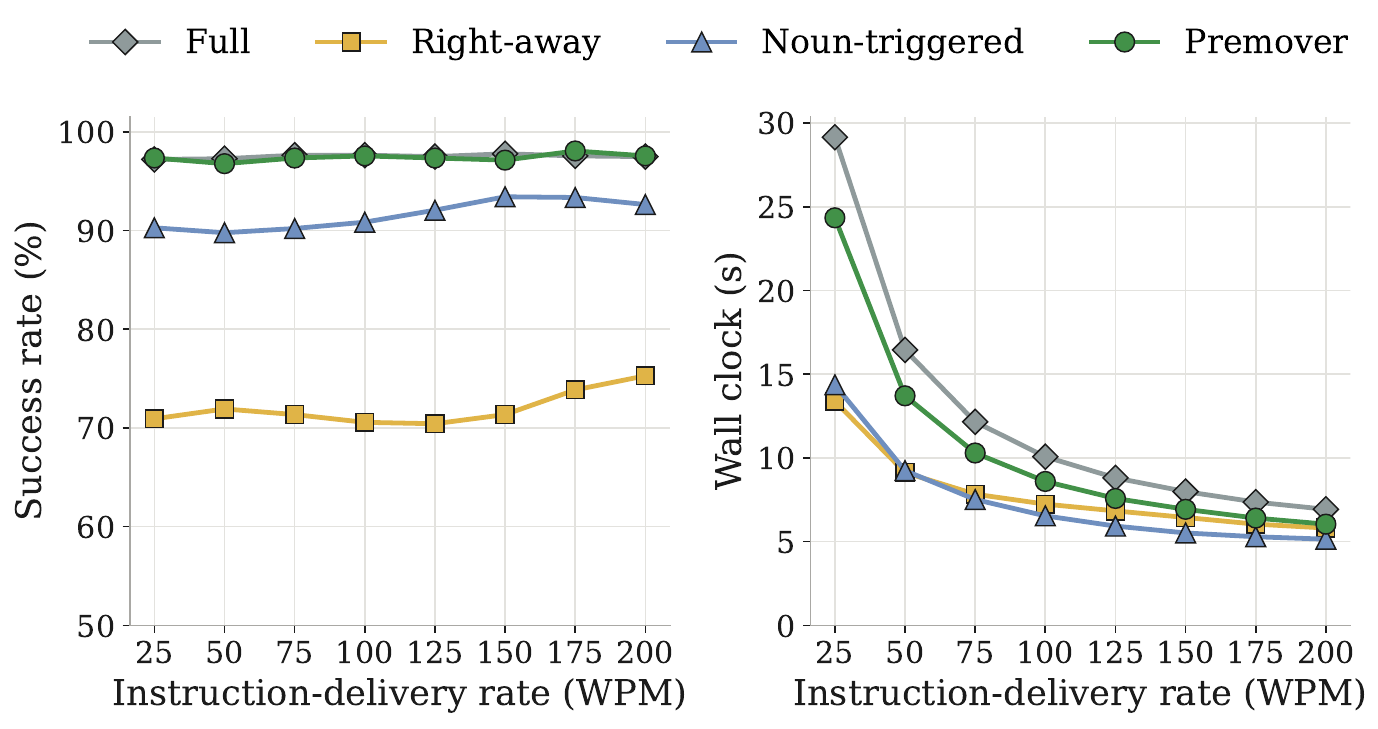}
    \caption{\textbf{Effect of instruction-delivery rate.} \method{} maintains high success rates and reduces wall-clock time across instruction-delivery rates.}
    \label{fig:abl-wpm}
    \vspace{-2.0em}
\end{wrapfigure}

We vary the instruction-delivery rate from $25$ to $200$ WPM at $25$-WPM intervals using $\pi_{0.5}$ on LIBERO, with streaming prefixes arriving according to each delivery rate. Wall-clock time includes both instruction delivery and task execution, accounting for any overlap between them.

\linkedref{Figure}{fig:abl-wpm} shows that \method{} consistently maintains Full-prompt-level success across all tested instruction-delivery rates, with success rates between $96.8$-$98.1\%$. At the same time, \method{} reduces wall-clock time relative to Full-prompt at every delivery rate, by $13$-$17\%$. These results show that \method{} preserves task performance while consistently reducing end-to-end interaction time across a wide range of instruction-delivery rates.

\subsubsection{Component Ablation}
\label{sec:abl_components}

\begingroup
\setlength{\intextsep}{0pt}
\setlength{\abovecaptionskip}{0pt}
\setlength{\belowcaptionskip}{2pt}

\begin{wraptable}{r}{0.46\textwidth}

\centering
\footnotesize
\setlength{\tabcolsep}{2.5pt}
\renewcommand{\arraystretch}{1.1}
\caption{\textbf{Component ablation.} Success rate (SR; \%) and wall-clock time (Wall; s) as the action readiness gate and vision-language focus map are added to the streaming baseline.}
\label{tab:abl-components}

\begin{tabularx}{\linewidth}{@{}ccc *{2}{>{\centering\arraybackslash}X}@{}}
\toprule
Streaming & Ready & Focus & \textbf{SR} $\uparrow$ & \textbf{Wall} $\downarrow$ \\
\midrule
\checkmark &            &            & 58.4 & 8.60 \\
\checkmark & \checkmark &            & 92.2 & 8.01 \\
\rowcolor{green!10}
\checkmark & \checkmark & \checkmark & 93.8 & 8.12 \\
\bottomrule
\end{tabularx}

\vspace{-6pt}
\end{wraptable}

To isolate the contribution of each component, we ablate the action readiness gate and vision-language focus map using GR00T-N1.7 on LIBERO at the speaking rate. \linkedref{Table}{tab:abl-components} summarizes the resulting changes in success rate and wall-clock time. Starting from the streaming-only baseline, adding the action readiness gate raises success from $58.4\%$ to $92.2\%$, accounting for most of the improvement over Right-away.

Adding the vision-language focus map further increases success to $93.8\%$. Overall, the results show that the action readiness gate contributes most of the improvement in success, while the focus map provides an additional gain that recovers nearly all of the remaining gap to Full-prompt.

\par
\endgroup

\subsubsection{Fixed-Fraction Execution Baselines}
\label{sec:abl_fixed_fraction}

\begingroup
\setlength{\intextsep}{0pt}
\setlength{\abovecaptionskip}{0pt}
\setlength{\belowcaptionskip}{2pt}
\begin{wraptable}{r}{0.48\textwidth}
\centering
\footnotesize
\setlength{\tabcolsep}{4pt}
\renewcommand{\arraystretch}{1.1}
\caption{\textbf{Fixed-fraction execution baselines.} Success rate (SR; \%) and wall-clock time (Wall; s) when execution begins after a fixed fraction of the instruction has been received.}
\label{tab:abl-fixed-fraction}
\begin{tabularx}{\linewidth}{
l
>{\centering\arraybackslash}X
>{\centering\arraybackslash}X
}
\toprule
\textbf{Setting} & \textbf{SR} $\uparrow$ & \textbf{Wall} $\downarrow$ \\
\midrule
Fixed $0\%$ (Right-away)    & 58.4 & 8.60 \\
Fixed $25\%$                & 69.3 & 8.07 \\
Fixed $50\%$                & 85.8 & 7.79 \\
Fixed $75\%$                & 93.6 & 8.30 \\
Fixed $100\%$ (Full-prompt) & 93.7 & 8.63 \\
\rowcolor{green!10}
\textbf{\method{}}          & 93.8 & 8.12 \\
\bottomrule
\end{tabularx}
\end{wraptable}

\linkedref{Table}{tab:abl-fixed-fraction} shows the fixed-fraction results using GR00T-N1.7 on LIBERO at the speaking rate. The fixed baselines start execution after receiving $0\%$, $25\%$, $50\%$, $75\%$, or $100\%$ of the instruction, where Fixed $0\%$ and $100\%$ correspond to \emph{Right-away} and \emph{Full-prompt}, respectively. Because these baselines depend on the fraction of the complete instruction, this controlled comparison uses the known full instruction to determine each execution point.

Starting earlier allows more of the instruction-delivery interval to overlap with execution, but can substantially reduce success. A fixed-fraction strategy also requires knowing the full instruction length in advance. \method{}, however, determines when the current streaming prefix contains sufficient information to begin execution, and thus reduces wall-clock time while maintaining near-Full-prompt success.

\par
\endgroup

\section{Discussion and Future Work}
\label{sec:discussion}

The benefit of \method{} depends on how much of the instruction-delivery interval can be safely overlapped with robot execution. When the information needed to identify the target and support the initial action appears early in the instruction, \method{} has more opportunity to begin execution before the complete instruction arrives. In contrast, the latency benefit can be smaller when the relevant information appears late, when instructions are delivered quickly, or when little execution can be completed before the remaining instruction arrives. Overall, the achievable latency reduction depends on the temporal structure of the instruction and the point at which the streaming prefix contains sufficient information to support execution.

Our current setting assumes that the intended task remains fixed as the instruction is revealed. We do not consider corrections or changes to previously specified targets or destinations after execution has begun, which would require the robot to reconsider actions already taken under an earlier prefix. Extending streaming execution to handle such revisions is an important direction for future work. \method{} also relies on offline supervision for the vision-language focus map and action readiness gate, although this supervision is generated automatically using a frozen segmentation model and a language model rather than manual annotation.

\section{Conclusion}

We introduced \method{}, a lightweight module that enables frozen VLA policies to begin execution before the complete instruction arrives. \method{} uses a vision-language focus map to identify where the current streaming prefix refers in the visual scene and an action readiness gate to determine when the prefix contains sufficient information to begin execution. At the speaking rate, \method{} reduces mean wall-clock time by $8.3\%$ on LIBERO with $\pi_{0.5}$ while maintaining comparable success ($97.6\%$ versus $97.9\%$), and by $17.9\%$ on the SO-101 robot while maintaining task success rates ($86.0\%$ versus $81.0\%$). These results show that overlapping instruction delivery with robot execution can reduce end-to-end interaction time while preserving task performance, without retraining the VLA backbone.

\begingroup
\raggedright
\bibliography{reference}
\bibliographystyle{plainnat}
\endgroup

\appendix
\section{Implementation Details}
\label{app:impl}

\subsection{Training Setup}
\label{app:training_setup}

\paragraph{Trainable parameters and optimization.}
The trainable parameters are the image and language projection heads
$f_{\mathrm{img}}$ and $f_{\mathrm{lang}}$,
the shared layer-mixing logits $\gamma$,
the readiness predictor $f_{\mathrm{ready}}$,
and the readiness threshold $\tau$.
The VLA backbone remains frozen throughout training.

The projection heads are two-layer GELU MLPs with hidden width $512$ and output width $256$, and the readiness predictor $f_{\mathrm{ready}}$ is a two-layer MLP with hidden width $256$.
The similarity scale in Eq.~\eqref{eq:focus-prob} is fixed to $\beta=6$.
The final layer of $f_{\mathrm{ready}}$ is initialized to zero, corresponding to an initial readiness score of $0.5$, and $\tau$ is initialized to $0.1$.

We optimize all trainable parameters with AdamW using a learning rate of $10^{-4}$, weight decay of $10^{-4}$, and global gradient-norm clipping at $1$.
The focus-map loss $\mathcal{L}_{\mathrm{focus}}$, region-contrastive loss $\mathcal{L}_{\mathrm{contrast}}$, and readiness loss $\mathcal{L}_{\mathrm{ready}}$ in Eq.~\eqref{eq:total-loss} each have unit weight.

\paragraph{Training data and prefix sampling.}
For each frozen backbone-benchmark pair, we train a separate set of heads across the selected tasks.
Demonstrations 0--9 are used for training, with up to $16$ sampled frames per demonstration and up to $8$ streaming prefixes per frame.
Prefix sampling includes the complete instruction and, when available, the last not-ready and first ready prefixes.
Training makes one pass over the sampled examples, with one optimizer update per successfully loaded example.

Demonstrations 10--14 are used for validation, and initial states 15--49 are reserved for rollout evaluation.
The evaluated backbones are $\pi_{0.5}$~\citep{black2025pi05} and GR00T-N1.7~\citep{nvidia2025groot,nvidia2026gr00tn17}.

\subsection{Vision-Language Focus Map Supervision}
\label{app:focus_supervision}

\paragraph{Patch-level focus supervision.}
Simulator instance masks provide offline supervision for the vision-language focus map.
Each mask is projected onto the corresponding camera patch grid, and patches with instance coverage greater than $0.02$ are included in the binary supervision mask.

The supervised region follows the task phase.
The target-object mask is used through the first positive gripper-close command in the demonstration and the following three frames, after which supervision switches to the destination mask.
Only regions identified by the current streaming prefix are included.
If the current streaming prefix has not yet identified the region for the corresponding task phase, the supervision mask is all zero.

Let
$m^\star\in\{0,1\}^{N}$
denote the binary patch-level supervision mask.
Each label $m_i^\star$ indicates whether image patch $i$ belongs to the supervised target-object or destination region.
The focus-map loss is

\begin{equation}
\begin{aligned}
\mathcal{L}_{\mathrm{focus}}
&=
-\frac{
\sum_{i=1}^{N}
v_i
\left[
m_i^\star\log p_i
+
(1-m_i^\star)\log(1-p_i)
\right]
}{
\max\left(\sum_{i=1}^{N}v_i,1\right)
},\\
v_i
&=
\begin{cases}
N_-/\max(N_+,1), & m_i^\star=1,\\
1, & m_i^\star=0,
\end{cases}
\end{aligned}
\label{eq:focus-loss}
\end{equation}

where
$N_+=\sum_i m_i^\star$
and
$N_-=N-N_+$.
All-zero masks are retained during training so that the focus map assigns low scores when the current streaming prefix has not yet identified a relevant region.

\paragraph{Region-contrastive supervision.}
The positive region is the union of the target-object and destination regions identified by the current streaming prefix.
Each negative region corresponds to another visible object and includes all patches with nonzero coverage by its segmentation mask.
A negative region is excluded when more than half of its patches overlap the positive region.

The region-contrastive loss encourages a higher mean focus score in prefix-relevant regions than in other visible regions.
Let $\mu_0$ denote the mean focus score in the positive region and $\mu_r$ the mean focus score in negative region $r$, with $J_-$ denoting the number of remaining negative regions.
We use

\begin{equation}
\mathcal{L}_{\mathrm{contrast}}
=
-\log
\frac{\exp(\mu_0/T_c)}
{\exp(\mu_0/T_c)+
\sum_{r=1}^{J_-}\eta_r\exp(\mu_r/T_c)},
\qquad
T_c=0.1.
\label{eq:region-contrast}
\end{equation}

For each negative region, we set
$\eta_r=4$ if its object is explicitly excluded by the current streaming prefix,
$\eta_r=2$ if the object has not yet been named,
and $\eta_r=1$ otherwise.
Explicit exclusion takes precedence when both conditions apply.
The loss is set to zero when the positive region is empty or no negative regions remain.

\subsection{Action Readiness Gate Details}
\label{app:readiness_details}

\paragraph{Readiness representation.}
The readiness predictor uses frozen backbone hidden states before the image and language projection heads.
At streaming-prefix update $u$, let $t$ denote the corresponding control step.

For $\pi_{0.5}$, we mean-pool the image patches and visible language tokens within each backbone layer, aggregate them using the shared layer weights $\omega_b$ from Eq.~\eqref{eq:focus-features}, normalize the image and language representations independently, and concatenate them:

\begin{equation}
x_u=
\left[
\operatorname{norm}\left(
\sum_{b=1}^{B}
\omega_b
\frac{1}{N}
\sum_{i=1}^{N}
h_{\mathrm{img},i}^{b}
\right);
\operatorname{norm}\left(
\sum_{b=1}^{B}
\omega_b
\frac{1}{L_t}
\sum_{j=1}^{L_t}
h_{\mathrm{lang},j}^{b}
\right)
\right].
\label{eq:readiness-features-pi05}
\end{equation}

Padding tokens are excluded from the language mean.

For GR00T-N1.7, we use the normalized layer-weighted hidden state of the last visible language token:

\begin{equation}
x_u=
\operatorname{norm}\left(
\sum_{b=1}^{B}
\omega_b h_{\mathrm{lang},L_t}^{b}
\right).
\label{eq:readiness-features-gr00t}
\end{equation}

Both representations are passed to the readiness predictor
$f_{\mathrm{ready}}$
in Eq.~\eqref{eq:readiness-score}.

\paragraph{Readiness supervision.}
We use Qwen3.5-2B to annotate an offline readiness cutoff from the complete instruction and scene information.
The cutoff is the earliest instruction word at which the streaming prefix contains sufficient information to identify the target and support the initial action.
When target identification requires a prepositional phrase, the cutoff is placed at the final word of that phrase.

Let $c$ denote the one-based index of the annotated readiness word.
At streaming-prefix update $u$, a prefix containing $k_t$ words receives the readiness label

\begin{equation}
y_u=\mathbf{1}[k_t\geq c].
\end{equation}

Prefixes before $c$ are labeled not ready, while prefixes at or beyond $c$ are labeled ready.

To jointly optimize the readiness score and learned threshold, we define

\begin{equation}
q_u^{\mathrm{ready}}
=
\sigma\left(
\frac{r_u-\tau}{0.1}
\right),
\end{equation}

and use the class-weighted binary cross-entropy loss

\begin{equation}
\mathcal{L}_{\mathrm{ready}}
=
-w_{y_u}
\left[
y_u\log q_u^{\mathrm{ready}}
+
(1-y_u)\log(1-q_u^{\mathrm{ready}})
\right],
\qquad
w_1=\frac{1}{2\rho},
\quad
w_0=\frac{1}{2(1-\rho)},
\label{eq:ready-loss}
\end{equation}

where $\rho$ is the fraction of ready prefixes and is clipped away from zero and one.

\paragraph{Execution start.}
The binary readiness decision $e_u$ is evaluated at streaming-prefix updates.
Execution begins when $e_u=1$ at two consecutive updates with distinct word counts.
Repeated evaluation of the same streaming prefix does not advance this condition.
We denote the resulting execution-start time by $t_{\mathrm{ready}}$.
Before $t_{\mathrm{ready}}$, the robot holds its pose.

\subsection{Simulation Evaluation Setup}
\label{app:simulation_setup}

\paragraph{Evaluation rollouts.}
The main simulation results are averaged over three evaluation seeds ($0$, $1$, and $2$), while all ablation studies use seed $0$.
For each evaluation seed, we run $35$ rollouts per task from initial states 15--49, corresponding to $350$ rollouts per LIBERO suite and $1400$ rollouts across the four suites.

Full-prompt, Right-away, Noun-triggered, and \method{} use the same seeded scene construction, initial states, and policy-noise reset procedure.
Comparisons match episodes by evaluation seed, suite, task, and initial state.

\paragraph{Computing resources.}
All experiments were conducted on $8$ NVIDIA H200 GPUs with 141\,GB of memory each. Each set of trainable heads was trained on a single GPU.

\section{Vision-Language Focus Map Details}
\label{app:focus-map}

\begin{figure}[H]
\centering
\includegraphics[width=\linewidth]{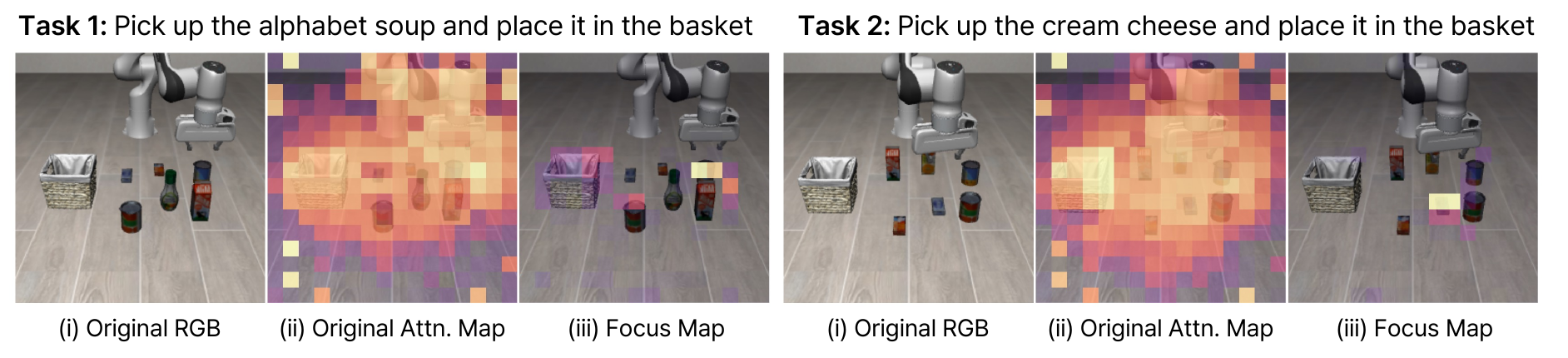}
\caption{\textbf{From dispersed attention to a focus map.} The learned vision-language focus map converts the frozen backbone's dispersed vision-language alignment into a localized relevance signal for the instruction referent. The panels show the original RGB observation, the frozen backbone's implicit alignment, and the learned focus map.}
\label{fig:dispersion}
\end{figure}

The vision-language focus map processes image patches from each camera in its own coordinate grid. Patches from the available camera views are concatenated to form the full patch representation, but corresponding grid indices in different views are not treated as the same physical location. Region components are therefore extracted separately for each camera before being ranked jointly across views. As a result, different views of the same physical object can produce separate region hypotheses.

The vision-language focus map computes focus scores independently using a sigmoid rather than a softmax over patches. Multiple regions can therefore receive high focus scores at the same time when the visible streaming prefix is compatible with several regions. This property is used by consensus-guided execution to construct multiple region-conditioned action chunks.

The action readiness gate uses backbone features aggregated with the same shared layer weights, with the backbone-specific representations given in \linkedref{Appendix}{app:impl}.

\section{Consensus-Guided Execution Details}
\label{app:consensus_details}

\begin{figure*}[ht]
\centering
\includegraphics[width=\linewidth]{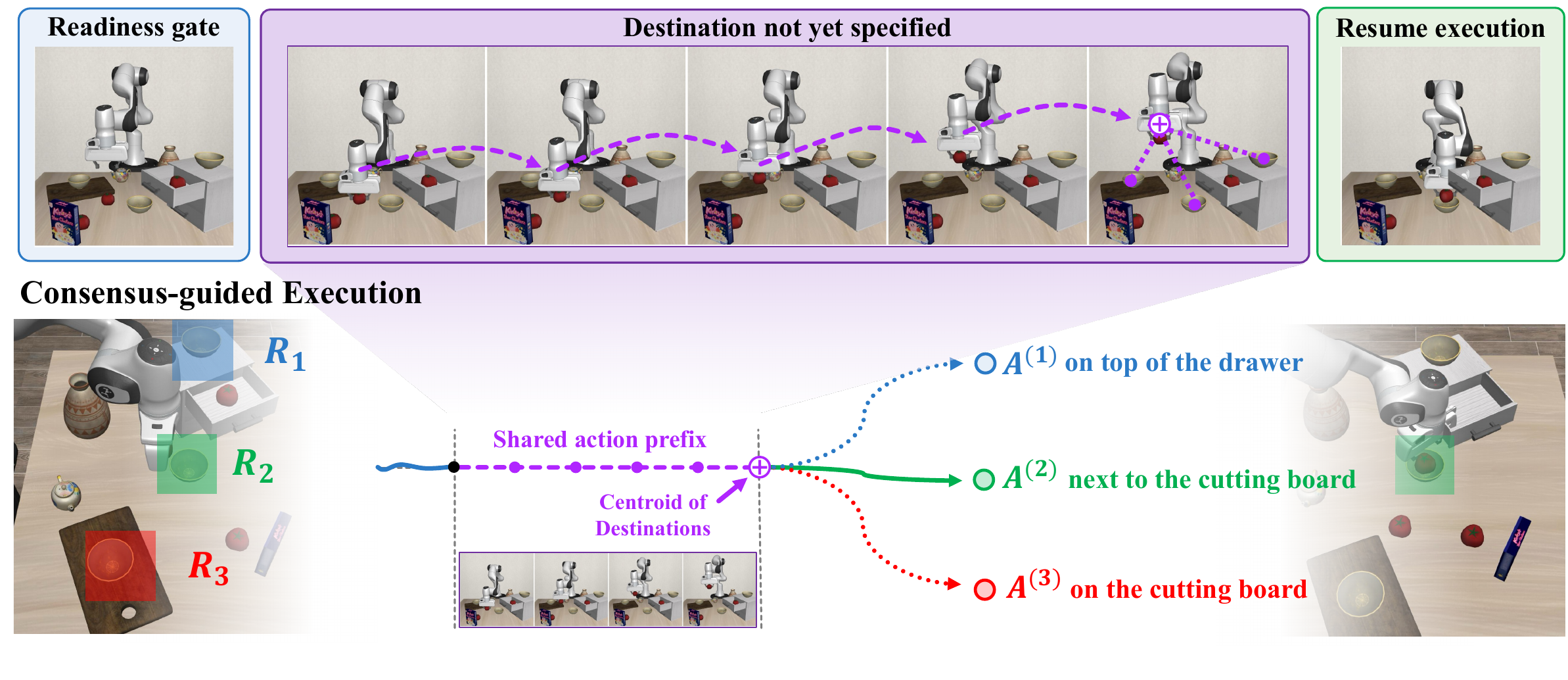}
\caption{\textbf{Consensus-guided execution under destination ambiguity.} After identifying and grasping the target object, \method{} executes the shared action prefix of the region-conditioned action chunks, defers execution when they diverge, and resumes default execution once the destination is specified. Dashed trajectories denote alternative predictions that are not executed.}
\label{fig:hedge_commit}
\end{figure*}

This section specifies region-hypothesis construction, the shared action prefix, and the conditions for resuming default execution used by consensus-guided execution in \linkedref{Section}{sec:method_execution}.

\subsection{Region Hypotheses and Convergence}
\label{app:consensus_regions}

\paragraph{Region hypothesis construction.}
We extract a ranked list of region hypotheses $\mathcal{R}=(R_1,\ldots,R_J)$ from the cached focus map. For each camera, we retain patches whose focus score is at least $5\%$ of the maximum focus score in that camera's focus map and extract four-connected components. Components from all camera views are ranked by their summed focus scores.

\paragraph{Region-conditioned action chunks.}
Let $m_i^{(d)}$ indicate whether patch $i$ belongs to region hypothesis $R_d$, where $d=1,\ldots,J$. We construct a soft visual weighting
\begin{equation}
w_i^{(d)}=\alpha+(1-\alpha)m_i^{(d)}.
\label{eq:focus-gain}
\end{equation}
The selected region retains weight one, while visual input outside the region is attenuated by $\alpha$. We use $\alpha=0.4$ for LIBERO and $\alpha=0.8$ for VLA-Arena.

For a fixed observation and streaming prefix, the frozen policy produces the default action chunk $A^{(0)}$ and one region-conditioned action chunk $A^{(d)}$ for each region hypothesis. Predictions within the same planning decision share the same sampling noise or seed. Each region-conditioned action chunk represents the policy's response with that region emphasized.

\paragraph{Cached focus map.}
Region hypotheses are extracted from the focus map cached at the previous fresh policy query. At the current query, the frozen backbone produces the default action chunk $A^{(0)}$ and features for the current focus map and readiness score. Region-conditioned action chunks use the current observation and streaming prefix with region masks from the cached map; the current focus map is then retained for the next fresh query. While waiting for readiness or for a new word after a shared-action-prefix decision, held control steps reuse the cached state without another policy query. The cache is empty at the beginning of an episode and is cleared at every reset.

\paragraph{Region convergence.}
Region hypotheses are ranked by their summed focus scores, while convergence is evaluated using their mean focus scores. We apply a softmax with temperature $0.1$ to the mean scores and require three conditions: a top-two probability margin of at least $0.5$, Jensen-Shannon divergence from the previous region distribution of at most $0.02$, and intersection-over-union of at least $0.1$ between the current and previous top regions. All conditions must hold at two consecutive updates with distinct word counts. Once reached, convergence remains satisfied for the rest of the episode, and its history is retained after the gripper-close command. A single remaining region hypothesis must also satisfy the temporal stability conditions before default execution resumes.

\subsection{Executing a Shared Action Prefix}
\label{app:consensus_bursts}

After the readiness condition is met, the robot executes default action chunks with the default number of actions executed from each predicted action chunk $P_{\mathrm{exec}}$ between replanning decisions. In the evaluated implementation, consensus-guided execution is enabled after the first executed gripper-close command, provided that the instruction is incomplete and the region hypotheses have not converged. Any remaining queued actions finish before consensus-guided execution is applied at the next planning decision. At an eligible decision, \method{} first checks the conditions for resuming default execution (\linkedref{Appendix}{app:consensus_joint}); otherwise, it executes only the shared action prefix defined below or waits for additional words.

\paragraph{Shared action prefix.}
For a set of predicted action chunks $\mathcal{A}$, let $M_{\mathrm{common}}$ denote the minimum available chunk length, equal to $M$ when all chunks have the prediction horizon defined in \linkedref{Section}{sec:methods}. For each chunk $A^{(d)}\in\mathcal{A}$, define cumulative translation and gripper command at action step $s$ as
\begin{equation}
x^{(d)}(s)=\sum_{v=1}^{s}a^{(d)}_{v,1:3},
\qquad
g^{(d)}(s)=\operatorname{grip}(a^{(d)}_s).
\end{equation}
The length of the shared action prefix, $S(\mathcal{A})$, is the longest initial interval over which all predicted trajectories remain within the translation and gripper tolerances:

\begin{equation}
\begin{aligned}
S(\mathcal{A})=\max\bigl\{m\in\{0,\ldots,M_{\mathrm{common}}\}:\;&
\max_{d,d'}\lVert x^{(d)}(s)-x^{(d')}(s)\rVert_2\leq\epsilon_x,\\[-1mm]
&\max_d g^{(d)}(s)-\min_d g^{(d)}(s)\leq\epsilon_g,\quad \forall s\leq m\bigr\}.
\end{aligned}
\label{eq:premove_horizon}
\end{equation}

We use $\epsilon_x=0.15$ in action-delta units and $\epsilon_g=0.5$. The comparison evaluates cumulative translation and gripper commands along the predicted trajectories.

\paragraph{Execution length.}
For $J\geq2$ region hypotheses, when none of the default-execution resumption conditions is satisfied, we compute the execution length as in Eq.~\eqref{eq:premove_burst}:
\begin{equation}
S_{\mathrm{cons}}
=
S\left(\{A^{(1)},\ldots,A^{(J)}\}\right),
\qquad
m_{\mathrm{exec}}^{\mathrm{cons}}
=
\min\left(S_{\mathrm{cons}},C,P_{\mathrm{exec}}\right).
\end{equation}
We use $C=3$ and take the first $m_{\mathrm{exec}}^{\mathrm{cons}}$ actions from $A^{(1)}$, the chunk conditioned on the highest-ranked region hypothesis $R_1$. Each streaming prefix permits at most one execution of a shared action prefix. The robot then holds its pose and preserves the last gripper command until another word arrives, skipping further policy and region-conditioned forwards for the unchanged prefix. If $S_{\mathrm{cons}}=0$, the robot remains still and waits for the next word before checking again. With fewer than two region hypotheses, no shared action prefix is executed at that decision.

\subsection{Resuming default Execution}
\label{app:consensus_joint}

Default execution resumes when region convergence, action agreement, or instruction completion is reached.

\paragraph{Action agreement.}
The action-agreement test is first performed at the first planning decision after the first executed gripper-close command for which the instruction is incomplete and the region hypotheses have not converged. It requires at least two valid region-conditioned action chunks and compares them with the default action chunk from the same decision:
\begin{equation}
S\!\left(\{A^{(0)},A^{(1)},\ldots,A^{(J)}\}\right)
>
T_{\mathrm{resume}}.
\end{equation}
The comparison uses the full common prediction horizon without the limit $C$ or the execution horizon $P_{\mathrm{exec}}$ and reuses the available chunks without an additional policy forward. If the test fails or fewer than two valid region-conditioned action chunks are available, the robot continues consensus-guided execution. No test is repeated during the unchanged-prefix wait following a shared-action-prefix decision.

We use $T_{\mathrm{resume}}=6$ for $\pi_{0.5}$ and $T_{\mathrm{resume}}=10$ for GR00T-N1.7. The prediction horizon is determined by the loaded backbone, and the test uses the full horizon common to the compared chunks.

\section{Instruction-Delivery Rates and Streaming Protocols}
\label{app:input_rate}
\label{app:streaming}

\paragraph{Typing rate.}
We derive the typing-rate condition from the public keystroke dataset of~\citet{dhakal2018observations}. WPM is computed using the standard five-character definition of a word. We first average the 15 sentence-level measurements available for each participant and then take the median across participants. The resulting participant-level median is $49.83$~WPM; we therefore use the rounded value of $50$~WPM as the representative typing rate.

\paragraph{Speaking rate.}
We derive the speaking-rate condition from Fair-Speech~\citep{veliche2024towards}, which contains approximately $26.5$K voice-command utterances from 593 speakers. Because the original audio durations include leading and trailing silence, we apply voice activity detection and compute WPM over active-speech spans. Of $26{,}470$ utterances, $26{,}466$ yield valid active-speech spans. The active-speech estimate has a median of $149.7$~WPM; we use the rounded value of $150$~WPM as the representative speaking rate.

\paragraph{Real-time instruction delivery.}
In simulation, instruction words arrive according to elapsed host time. The scheduling clock $t_{\mathrm{sched}}$ starts at the first control tick after the policy-noise and environment resets, before the settling steps, and advances using the monotonic timer \texttt{time.perf\_counter}. A word with recorded arrival time $s$ becomes visible at the first prefix check for which $t_{\mathrm{sched}}\geq s$. Policy inference, prediction of region-conditioned action chunks, holding, settling, and action execution all consume elapsed time on this clock. When diagnostic state audits are enabled, their measured duration is excluded from the input-arrival clock. Words that arrive during a policy forward or a queued action sequence are incorporated at a subsequent policy query.

\paragraph{Calibration, idle pacing, and warm-up.}
The calibration procedure measures average host time per rollout step on validation initial states 10 and 11, excluding the first episode for compilation and cache warm-up. All methods in a comparison reuse the same calibration record to set the idle control period $\Delta t$ and rollout step budget. A holding or settling tick lasts at least $\Delta t$, with an explicit wait when simulator computation finishes earlier. Active action steps take their measured computation time without this additional pacing. The calibration period therefore controls idle pacing and the step budget; instruction arrival itself uses elapsed host time rather than $n\Delta t$. Before timed evaluation, the common warm-up procedure uses validation initial state 10 to query the incoming prefixes and region-conditioned action chunks. Warm-up and model loading are outside the reported timing interval.

\paragraph{Incremental-text protocol.}
For uniform text streams at a rate $v$ WPM, the number of visible instruction words at control step $t$ is
\begin{equation}
k_t
=
\min\left(
W,\,
1+\left\lfloor\frac{t_{\mathrm{sched}}\,v}{60}\right\rfloor
\right).
\label{eq:streaming_clock}
\end{equation}
Here, $W$ is the total number of instruction words. At the typing rate of $50$~WPM, one word arrives every $1.2$~s, with the first word visible at scheduling time zero. The scheduled interval from the first word to instruction completion is $(W-1)60/v$ seconds; each arrival becomes visible at the next prefix check. The tokenizer processes the currently visible words at each policy query. The instruction-delivery-rate ablation in \linkedref{Section}{sec:abl_rate} also uses incrementally revealed text, with $v\in\{25,50,75,100,125,150,175,200\}$~WPM. Its $150$-WPM condition uses this uniform text schedule, whereas the speaking-rate results use the ASR replay protocol below.

\paragraph{Speech input and replay.}
We use XTTS-v2 for speech synthesis and faster-whisper large-v3 for transcription. At the speaking rate of $150$~WPM, simulation replays recorded ASR word-arrival timestamps against the elapsed host-time clock $t_{\mathrm{sched}}$. The recorded schedule preserves arrival delays and bursty emissions rather than assuming a uniform $0.4$-s interval between words. Speech synthesis and ASR are performed before replay, and the policy receives the prefix available at the current scheduling time. The real-robot results in \linkedref{Section}{sec:real_results} use speech and online transcription at the speaking rate; their wall-clock measurements include audio delivery and ASR latency.

\paragraph{Wall-clock measurement.}
\label{app:eval_metrics}
For the real-time simulation evaluation, Wall is the recorded rollout time \texttt{rollout\_wall\_s}. The raw episode timer \texttt{wall\_s} uses \texttt{time.perf\_counter}, starts before the policy-noise and environment resets, and stops after the episode loop terminates. The reported metric subtracts the measured reset and settling-computation costs:
\begin{equation}
\mathrm{Wall}
=t_{\mathrm{episode}}-t_{\mathrm{reset}}-t_{\mathrm{noise\ reset}}-t_{\mathrm{settle\ compute}}.
\label{eq:rollout-wall}
\end{equation}
It includes policy inference, additional region-conditioned forwards, readiness and focus computation, execution decisions, and simulator stepping during holding and action execution. Idle-pacing waits, including those during settling, remain included. Model loading, task-scene construction, and separately executed warm-up episodes are outside the raw episode timer.

Wall is averaged over all evaluated episodes, including unsuccessful episodes. It is measured directly, without converting the step count using $\Delta t$ or adding instruction-delivery time afterward. Relative Wall divides a method's mean by the Full-prompt mean under the same delivery condition, timing definition, and task coverage. Projection-head profiling is reported separately in \linkedref{Appendix}{app:focus-head-overhead}.

\paragraph{Input-waiting and action-time breakdown.}
\label{app:input_breakdown}
\linkedref{Figure}{fig:instructionT} reports measured input waiting and robot action times for Full-prompt with $\pi_{0.5}$ on LIBERO at $150$~WPM, using $4200$ episodes across three seeds.
Input waiting accounts for $65.9\%$, $57.5\%$, $50.0\%$, and $45.9\%$ of the wall-clock time on Spatial, Object, Goal, and Long-10, respectively. The overall share is $54.6\%$.

\section{Full Simulation Results}
\label{app:full_results}

This section reports the complete simulation results for the remaining backbone-benchmark combinations across both instruction-delivery rates.

\begin{table*}[ht]
\centering
\scriptsize
\setlength{\tabcolsep}{1.8pt}
\renewcommand{\arraystretch}{1.05}
\caption{\textbf{Main results on VLA-Arena with $\pi_{0.5}$.}
Success rate (SR; \%) and wall-clock time (Wall; s) across the four VLA-Arena suites at both instruction-delivery rates. Entries are means over three evaluation seeds, with seed-level standard deviations reported for both SR and Wall.}
\label{tab:vla_arena_pi05}

\begin{tabularx}{\textwidth}{@{}l *{10}{>{\centering\arraybackslash}X}@{}}
\toprule
& \multicolumn{2}{c}{\textbf{Safety}}
& \multicolumn{2}{c}{\textbf{Distractor}}
& \multicolumn{2}{c}{\textbf{Extrapolation}}
& \multicolumn{2}{c}{\textbf{Long Horizon}}
& \multicolumn{2}{c}{\textbf{Mean}} \\
\cmidrule(lr){2-3}
\cmidrule(lr){4-5}
\cmidrule(lr){6-7}
\cmidrule(lr){8-9}
\cmidrule(lr){10-11}

\textbf{Method}
& SR $\uparrow$ & Wall $\downarrow$
& SR $\uparrow$ & Wall $\downarrow$
& SR $\uparrow$ & Wall $\downarrow$
& SR $\uparrow$ & Wall $\downarrow$
& SR $\uparrow$ & Wall $\downarrow$ \\
\midrule

\multicolumn{11}{l}{\textit{Typing rate}} \\
Full-prompt
& 57.2{\tiny$\pm$1.0} & 18.45{\tiny$\pm$0.06}
& 78.7{\tiny$\pm$3.9} & 16.13{\tiny$\pm$0.15}
& 54.5{\tiny$\pm$1.3} & 24.40{\tiny$\pm$0.10}
& 82.9{\tiny$\pm$2.8} & 17.22{\tiny$\pm$0.17}
& 64.4{\tiny$\pm$0.2} & 19.35{\tiny$\pm$0.01} \\

Right-away
& 59.8{\tiny$\pm$1.1} & 10.56{\tiny$\pm$0.23}
& 76.3{\tiny$\pm$2.3} & 7.26{\tiny$\pm$0.36}
& 17.1{\tiny$\pm$0.6} & 24.53{\tiny$\pm$0.06}
& 79.4{\tiny$\pm$2.1} & 8.54{\tiny$\pm$0.43}
& 55.2{\tiny$\pm$0.3} & 13.16{\tiny$\pm$0.05} \\

Noun-triggered
& 60.3{\tiny$\pm$1.5} & 12.33{\tiny$\pm$0.21}
& 77.5{\tiny$\pm$1.7} & 9.14{\tiny$\pm$0.29}
& 35.7{\tiny$\pm$0.3} & 22.00{\tiny$\pm$0.11}
& 84.8{\tiny$\pm$1.9} & 10.29{\tiny$\pm$0.24}
& 61.1{\tiny$\pm$0.7} & 13.88{\tiny$\pm$0.08} \\

\rowcolor{green!10}
\textbf{\method{}}
& 61.2{\tiny$\pm$1.4} & 13.40{\tiny$\pm$0.22}
& 80.3{\tiny$\pm$0.3} & 10.61{\tiny$\pm$0.03}
& 41.6{\tiny$\pm$1.3} & 21.41{\tiny$\pm$0.31}
& 80.3{\tiny$\pm$4.0} & 12.57{\tiny$\pm$0.54}
& 62.7{\tiny$\pm$0.8} & 14.80{\tiny$\pm$0.06} \\

\midrule

\multicolumn{11}{l}{\textit{Speaking rate}} \\
Full-prompt
& 57.1{\tiny$\pm$1.6} & 9.71{\tiny$\pm$0.08}
& 77.8{\tiny$\pm$3.8} & 8.11{\tiny$\pm$0.18}
& 55.2{\tiny$\pm$1.7} & 13.46{\tiny$\pm$0.16}
& 82.1{\tiny$\pm$2.3} & 9.66{\tiny$\pm$0.14}
& 64.3{\tiny$\pm$0.8} & 10.37{\tiny$\pm$0.05} \\

Right-away
& 61.4{\tiny$\pm$1.9} & 7.57{\tiny$\pm$0.21}
& 78.8{\tiny$\pm$1.7} & 6.19{\tiny$\pm$0.11}
& 32.1{\tiny$\pm$1.4} & 14.04{\tiny$\pm$0.18}
& 83.3{\tiny$\pm$0.2} & 7.38{\tiny$\pm$0.07}
& 60.7{\tiny$\pm$0.7} & 8.93{\tiny$\pm$0.06} \\

Noun-triggered
& 60.2{\tiny$\pm$0.7} & 8.00{\tiny$\pm$0.06}
& 78.0{\tiny$\pm$2.1} & 6.53{\tiny$\pm$0.16}
& 44.7{\tiny$\pm$1.4} & 13.08{\tiny$\pm$0.14}
& 85.8{\tiny$\pm$2.2} & 7.67{\tiny$\pm$0.27}
& 63.5{\tiny$\pm$0.7} & 8.97{\tiny$\pm$0.04} \\

\rowcolor{green!10}
\textbf{\method{}}
& 58.0{\tiny$\pm$1.6} & 8.75{\tiny$\pm$0.09}
& 77.0{\tiny$\pm$1.9} & 7.81{\tiny$\pm$0.11}
& 49.9{\tiny$\pm$2.1} & 12.99{\tiny$\pm$0.18}
& 85.6{\tiny$\pm$1.4} & 8.56{\tiny$\pm$0.07}
& 63.7{\tiny$\pm$1.3} & 9.63{\tiny$\pm$0.08} \\

\bottomrule
\end{tabularx}
\end{table*}

\begin{table*}[ht]
\centering
\scriptsize
\setlength{\tabcolsep}{1.8pt}
\renewcommand{\arraystretch}{1.05}
\caption{\textbf{Main results on LIBERO with GR00T-N1.7.}
Success rate (SR; \%) and wall-clock time (Wall; s) across the four LIBERO suites at both instruction-delivery rates. Entries are means over three evaluation seeds, with seed-level standard deviations reported for both SR and Wall.}
\label{tab:libero_groot}

\begin{tabularx}{\textwidth}{@{}l *{10}{>{\centering\arraybackslash}X}@{}}
\toprule
& \multicolumn{2}{c}{\textbf{Spatial}}
& \multicolumn{2}{c}{\textbf{Object}}
& \multicolumn{2}{c}{\textbf{Goal}}
& \multicolumn{2}{c}{\textbf{Long-10}}
& \multicolumn{2}{c}{\textbf{Mean}} \\
\cmidrule(lr){2-3}
\cmidrule(lr){4-5}
\cmidrule(lr){6-7}
\cmidrule(lr){8-9}
\cmidrule(lr){10-11}

\textbf{Method}
& SR $\uparrow$ & Wall $\downarrow$
& SR $\uparrow$ & Wall $\downarrow$
& SR $\uparrow$ & Wall $\downarrow$
& SR $\uparrow$ & Wall $\downarrow$
& SR $\uparrow$ & Wall $\downarrow$ \\
\midrule

\multicolumn{11}{l}{\textit{Typing rate}} \\
Full-prompt
& 95.0{\tiny$\pm$1.2} & 20.63{\tiny$\pm$0.06}
& 98.5{\tiny$\pm$0.6} & 14.87{\tiny$\pm$0.02}
& 95.7{\tiny$\pm$0.8} & 10.37{\tiny$\pm$0.02}
& 87.6{\tiny$\pm$1.7} & 21.49{\tiny$\pm$0.09}
& 94.2{\tiny$\pm$0.4} & 16.84{\tiny$\pm$0.02} \\

Right-away
& 55.5{\tiny$\pm$1.6} & 13.11{\tiny$\pm$0.25}
& 50.3{\tiny$\pm$1.2} & 11.35{\tiny$\pm$0.17}
& 12.9{\tiny$\pm$2.0} & 14.06{\tiny$\pm$0.14}
& 60.0{\tiny$\pm$2.5} & 16.22{\tiny$\pm$0.29}
& 44.7{\tiny$\pm$0.9} & 13.68{\tiny$\pm$0.13} \\

Noun-triggered
& 55.4{\tiny$\pm$0.9} & 15.29{\tiny$\pm$0.16}
& 97.4{\tiny$\pm$1.0} & 8.02{\tiny$\pm$0.06}
& 52.2{\tiny$\pm$0.6} & 10.83{\tiny$\pm$0.06}
& 71.0{\tiny$\pm$0.7} & 16.25{\tiny$\pm$0.32}
& 69.0{\tiny$\pm$0.2} & 12.60{\tiny$\pm$0.06} \\

\rowcolor{green!10}
\textbf{\method{}}
& 95.1{\tiny$\pm$1.3} & 17.40{\tiny$\pm$0.14}
& 98.2{\tiny$\pm$0.4} & 12.33{\tiny$\pm$0.02}
& 96.8{\tiny$\pm$0.3} & 10.35{\tiny$\pm$0.02}
& 86.6{\tiny$\pm$1.3} & 19.85{\tiny$\pm$0.15}
& 94.2{\tiny$\pm$0.4} & 14.98{\tiny$\pm$0.03} \\

\midrule

\multicolumn{11}{l}{\textit{Speaking rate}} \\
Full-prompt
& 94.2{\tiny$\pm$0.9} & 9.21{\tiny$\pm$0.06}
& 97.9{\tiny$\pm$0.3} & 7.48{\tiny$\pm$0.02}
& 95.5{\tiny$\pm$0.7} & 5.71{\tiny$\pm$0.06}
& 88.2{\tiny$\pm$1.4} & 11.99{\tiny$\pm$0.07}
& 94.0{\tiny$\pm$0.3} & 8.60{\tiny$\pm$0.03} \\

Right-away
& 56.4{\tiny$\pm$0.9} & 7.97{\tiny$\pm$0.05}
& 49.6{\tiny$\pm$0.9} & 7.84{\tiny$\pm$0.08}
& 59.0{\tiny$\pm$1.2} & 6.97{\tiny$\pm$0.06}
& 65.7{\tiny$\pm$1.2} & 11.54{\tiny$\pm$0.03}
& 57.7{\tiny$\pm$0.9} & 8.58{\tiny$\pm$0.05} \\

Noun-triggered
& 58.4{\tiny$\pm$0.9} & 8.65{\tiny$\pm$0.04}
& 98.6{\tiny$\pm$0.3} & 5.84{\tiny$\pm$0.03}
& 76.9{\tiny$\pm$0.9} & 6.00{\tiny$\pm$0.06}
& 75.0{\tiny$\pm$2.9} & 11.31{\tiny$\pm$0.23}
& 77.2{\tiny$\pm$0.8} & 7.95{\tiny$\pm$0.06} \\

\rowcolor{green!10}
\textbf{\method{}}
& 94.3{\tiny$\pm$1.3} & 8.33{\tiny$\pm$0.04}
& 97.7{\tiny$\pm$1.1} & 6.60{\tiny$\pm$0.07}
& 95.6{\tiny$\pm$0.4} & 5.75{\tiny$\pm$0.03}
& 87.4{\tiny$\pm$1.0} & 11.61{\tiny$\pm$0.12}
& 93.8{\tiny$\pm$0.5} & 8.07{\tiny$\pm$0.05} \\

\bottomrule
\end{tabularx}
\end{table*}

\begin{table*}[ht]
\centering
\scriptsize
\setlength{\tabcolsep}{1.8pt}
\renewcommand{\arraystretch}{1.05}
\caption{\textbf{Main results on VLA-Arena with GR00T-N1.7.}
Success rate (SR; \%) and wall-clock time (Wall; s) across the four VLA-Arena suites at both instruction-delivery rates. Entries are means over three evaluation seeds, with seed-level standard deviations reported for both SR and Wall.}
\label{tab:vla_arena_groot}

\begin{tabularx}{\textwidth}{@{}l *{10}{>{\centering\arraybackslash}X}@{}}
\toprule
& \multicolumn{2}{c}{\textbf{Safety}}
& \multicolumn{2}{c}{\textbf{Distractor}}
& \multicolumn{2}{c}{\textbf{Extrapolation}}
& \multicolumn{2}{c}{\textbf{Long Horizon}}
& \multicolumn{2}{c}{\textbf{Mean}} \\
\cmidrule(lr){2-3}
\cmidrule(lr){4-5}
\cmidrule(lr){6-7}
\cmidrule(lr){8-9}
\cmidrule(lr){10-11}

\textbf{Method}
& SR $\uparrow$ & Wall $\downarrow$
& SR $\uparrow$ & Wall $\downarrow$
& SR $\uparrow$ & Wall $\downarrow$
& SR $\uparrow$ & Wall $\downarrow$
& SR $\uparrow$ & Wall $\downarrow$ \\
\midrule

\multicolumn{11}{l}{\textit{Typing rate}} \\
Full-prompt
& 41.4{\tiny$\pm$1.1} & 15.92{\tiny$\pm$0.01}
& 66.6{\tiny$\pm$2.3} & 14.30{\tiny$\pm$0.07}
& 42.9{\tiny$\pm$1.8} & 21.75{\tiny$\pm$0.08}
& 60.7{\tiny$\pm$4.4} & 16.13{\tiny$\pm$0.24}
& 49.2{\tiny$\pm$0.6} & 17.14{\tiny$\pm$0.02} \\

Right-away
& 39.4{\tiny$\pm$1.5} & 9.89{\tiny$\pm$0.14}
& 66.0{\tiny$\pm$3.0} & 6.87{\tiny$\pm$0.51}
& 11.3{\tiny$\pm$1.0} & 21.80{\tiny$\pm$0.21}
& 61.9{\tiny$\pm$1.6} & 7.64{\tiny$\pm$0.26}
& 40.6{\tiny$\pm$1.3} & 11.99{\tiny$\pm$0.20} \\

Noun-triggered
& 42.7{\tiny$\pm$0.3} & 11.31{\tiny$\pm$0.14}
& 69.0{\tiny$\pm$1.0} & 8.36{\tiny$\pm$0.12}
& 31.8{\tiny$\pm$2.3} & 19.44{\tiny$\pm$0.30}
& 60.4{\tiny$\pm$2.6} & 10.31{\tiny$\pm$0.26}
& 47.3{\tiny$\pm$0.3} & 12.68{\tiny$\pm$0.02} \\

\rowcolor{green!10}
\textbf{\method{}}
& 41.2{\tiny$\pm$0.4} & 13.14{\tiny$\pm$0.10}
& 64.1{\tiny$\pm$2.9} & 11.30{\tiny$\pm$0.16}
& 33.1{\tiny$\pm$1.2} & 19.99{\tiny$\pm$0.14}
& 61.3{\tiny$\pm$0.9} & 11.55{\tiny$\pm$0.09}
& 46.3{\tiny$\pm$1.0} & 14.28{\tiny$\pm$0.03} \\

\midrule

\multicolumn{11}{l}{\textit{Speaking rate}} \\
Full-prompt
& 41.0{\tiny$\pm$1.2} & 7.10{\tiny$\pm$0.05}
& 65.5{\tiny$\pm$2.3} & 6.22{\tiny$\pm$0.03}
& 41.8{\tiny$\pm$0.9} & 10.86{\tiny$\pm$0.11}
& 60.8{\tiny$\pm$1.7} & 8.47{\tiny$\pm$0.12}
& 48.6{\tiny$\pm$1.1} & 8.12{\tiny$\pm$0.05} \\

Right-away
& 38.0{\tiny$\pm$0.5} & 5.90{\tiny$\pm$0.00}
& 62.9{\tiny$\pm$1.2} & 4.80{\tiny$\pm$0.04}
& 15.1{\tiny$\pm$0.7} & 11.51{\tiny$\pm$0.08}
& 59.6{\tiny$\pm$2.1} & 6.78{\tiny$\pm$0.07}
& 40.0{\tiny$\pm$0.3} & 7.27{\tiny$\pm$0.01} \\

Noun-triggered
& 38.3{\tiny$\pm$1.0} & 6.03{\tiny$\pm$0.12}
& 66.9{\tiny$\pm$1.6} & 4.92{\tiny$\pm$0.06}
& 30.5{\tiny$\pm$2.0} & 10.61{\tiny$\pm$0.13}
& 57.8{\tiny$\pm$1.6} & 7.15{\tiny$\pm$0.20}
& 44.4{\tiny$\pm$0.8} & 7.18{\tiny$\pm$0.05} \\

\rowcolor{green!10}
\textbf{\method{}}
& 44.2{\tiny$\pm$2.0} & 6.53{\tiny$\pm$0.05}
& 65.7{\tiny$\pm$1.2} & 5.81{\tiny$\pm$0.04}
& 36.9{\tiny$\pm$0.5} & 10.66{\tiny$\pm$0.05}
& 58.1{\tiny$\pm$1.2} & 7.74{\tiny$\pm$0.04}
& 48.3{\tiny$\pm$0.7} & 7.64{\tiny$\pm$0.03} \\

\bottomrule
\end{tabularx}
\end{table*}

\section{Real-Robot Implementation Details}
\label{app:real_robot}

The real-robot results in \linkedref{Table}{tab:real_robot} use the frozen $\pi_{0.5}$ backbone on an SO-101 robot with speech and online transcription at the speaking rate ($150$~WPM). \linkedref{Figure}{fig:real_world_tasks} illustrates the task setups.

\paragraph{Task definitions and scoring.}
The five single-object tasks are Block in Drawer, Block in Cup, Cat in Cup, Cup in Cup, and Push Cup. Cat in Cup requires placing a gray cat-shaped object into a cup. All five use binary success. The three disambiguation tasks share the block-to-drawer goal and differ in the referring expression that identifies the target block: in front of the orange cat, on the cup, or next to the gray cat. They report partial success scores with equal subgoal weighting.

\paragraph{Single-object prompts.}
The five tasks use the following instructions:

\begin{itemize}
\setlength{\itemsep}{0pt}
\setlength{\parskip}{0pt}
    \item \textbf{Block$\rightarrow$Cup:} ``Pick up the block and place it in the cup on the right.''
    \item \textbf{Drawer:} ``Pick up the block and put it in the drawer then close it.''
    \item \textbf{Cat$\rightarrow$Cup:} ``Pick up the cat doll and place it in the cup on the right.''
    \item \textbf{Cup$\rightarrow$Cup:} ``Pick up one cup and place it inside the other cup.''
    \item \textbf{Push Cup:} ``Move the cup leftward by pushing it.''
\end{itemize}

\paragraph{Disambiguation prompts.}
The three instructions specify different spatial relations to identify the target block:
\begin{itemize}
\setlength{\itemsep}{0pt}
\setlength{\parskip}{0pt}
    \item \textbf{On Cup:} ``Pick up the block on the cup, place it in the drawer, and then close the drawer.''
    \item \textbf{Gray Cat:} ``Pick up the block next to the gray cat, place it in the drawer, and then close the drawer.''
    \item \textbf{Orange Cat:} ``Pick up the block in front of the orange cat, place it in the drawer, and then close the drawer.''
\end{itemize}

\paragraph{Evaluation protocol.}
For the single-object results, Wall is averaged over 20 measurements, including unsuccessful episodes. An episode terminates when the final goal state is reached or after a maximum of 900 policy steps. Real-robot Wall includes instruction delivery, online transcription, policy computation, and robot execution, accounting for their overlap. It follows the end-to-end timing definition in \linkedref{Section}{sec:exp_setup}; the simulator-specific reset and settling subtractions in Eq.~\eqref{eq:rollout-wall} do not apply to the robot.
For the disambiguation tasks, we assign 0.5 points for placing the target block in the drawer and 0.5 points for closing the drawer, giving a maximum score of 1.0 per episode. We report the mean episode score as a percentage.

\begin{figure}[t]
    \centering
    \includegraphics[width=\linewidth]{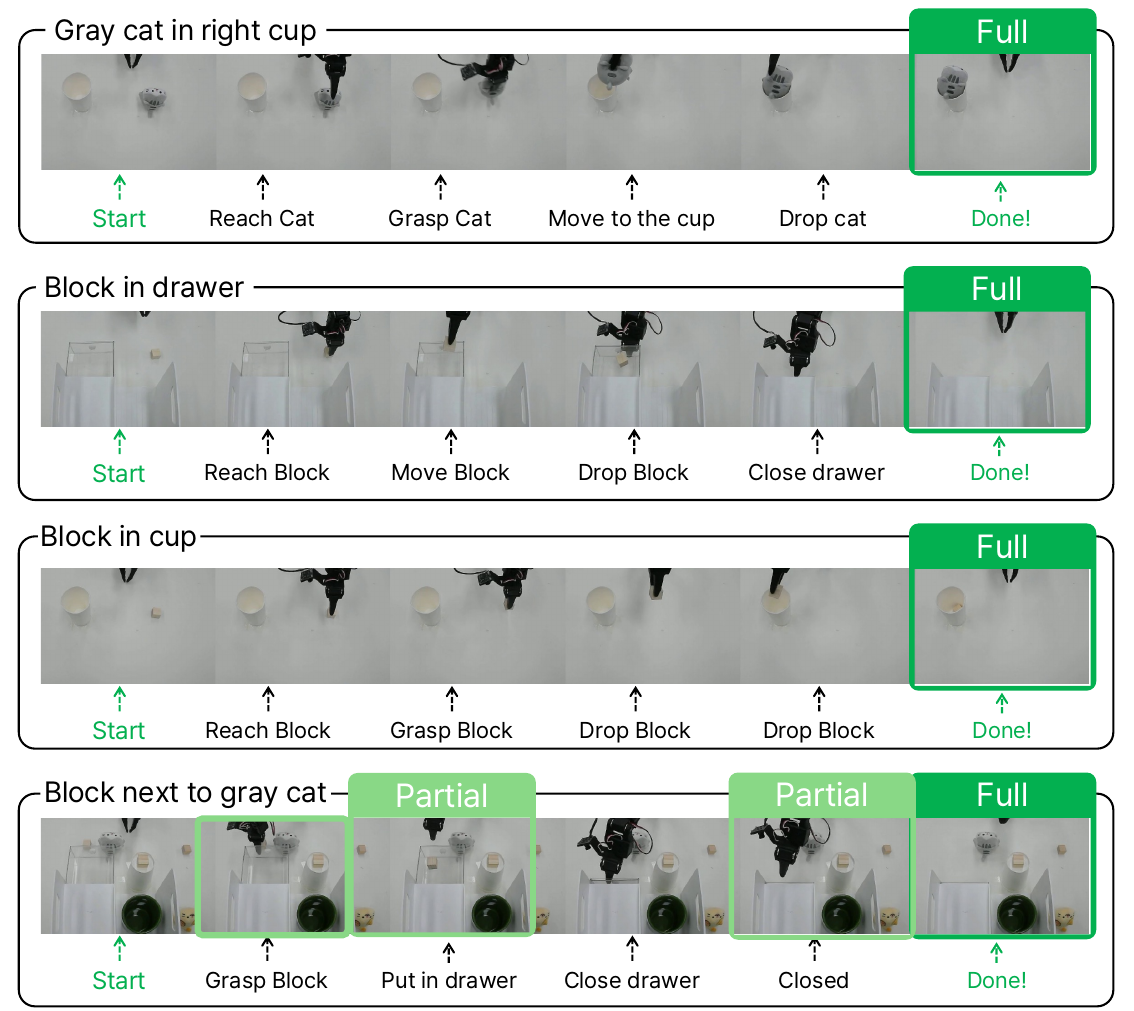}
    \caption{\textbf{Real-robot task setups on the SO-101.} The evaluation includes five single-object tasks and three disambiguation tasks with different spatial referring expressions.}
    \label{fig:real_world_tasks}
\end{figure}

\paragraph{Streaming implementation.}
Instructions are synthesized through TTS at the speaking rate and transcribed online. Audio capture uses sounddevice v0.5.5, streaming transcription uses faster-whisper v1.2.1, and host-side input events are handled using pynput v1.8.2. Each method receives the same audio and incremental transcript within a given task. Timing improvements are computed relative to the Full-prompt baseline evaluated under the same speaking-rate and transcription condition.

\paragraph{Data collection and supervision.}
We collect 50 teleoperated expert demonstrations for each task. Observations consist of synchronized third-person and wrist-mounted camera images together with the robot state. A frozen Segment Anything Model (SAM)~\citep{kirillov2023sam} generates offline object-mask pseudo-labels for training the vision-language focus map. The VLA backbone remains frozen. At evaluation, the VLA receives camera observations, robot state, and the streaming prefix.

\section{Sensitivity to the Focus Concentration Set Size}
\label{app:focus_k_sensitivity}

\begin{figure*}[ht]
   \centering
   \includegraphics[width=\textwidth]{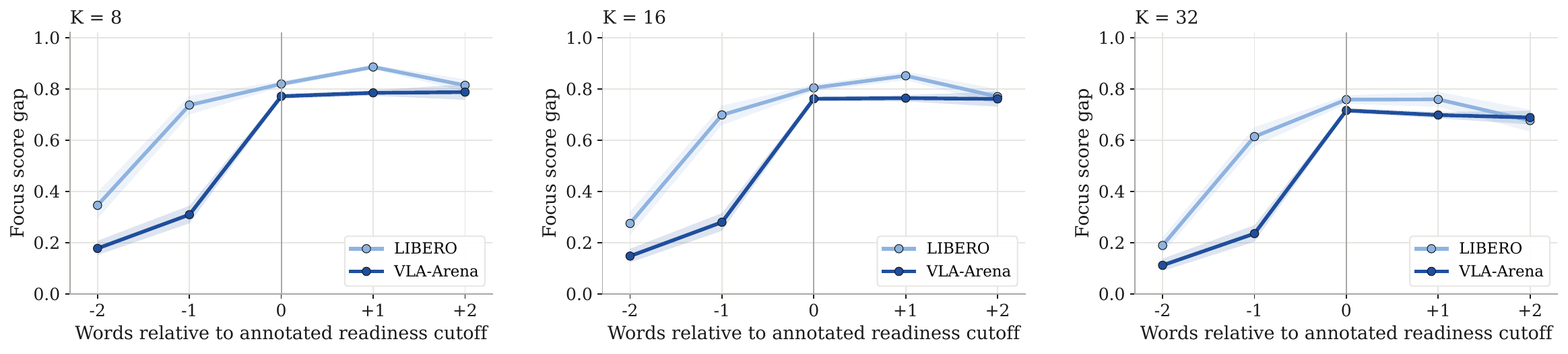}
   \caption{\textbf{Sensitivity of focus concentration to $K$.}
   Gap between the mean focus scores of the top-$K$ patches and the remaining patches for $K\in\{8,16,32\}$, averaged over $\pi_{0.5}$ and GR00T-N1.7. On both LIBERO and VLA-Arena, the gap increases sharply around the annotated readiness cutoff for all three values of $K$.}
   \label{fig:focus_k_sensitivity}
\end{figure*}

In the main analysis, we measure focus concentration using the gap between the mean focus score of the top-$K$ patches and that of the remaining patches, with $K=16$. To test whether the observed behavior depends on this choice, we repeat the analysis with $K\in\{8,16,32\}$.

As shown in \linkedref{Fig.}{fig:focus_k_sensitivity}, the same trend holds across all three values of $K$. Around the annotated readiness cutoff, the top-$K$ patches receive substantially higher focus scores, while the scores of the remaining patches stay comparatively low on both LIBERO and VLA-Arena. Increasing $K$ reduces the absolute top-$K$ mean, as more lower-scoring patches are included, but does not change the overall behavior. These results indicate that the concentration observed in the main analysis is not specific to the choice of $K=16$.

\paragraph{Joint-displacement metric.}
For the real-robot analysis in \linkedref{Fig.}{fig:analysis}(c), we measure physical motion relative to the robot's starting configuration. At each time $t$, we define the maximum joint displacement as
\[
d(t) = \max_j \left| q_j(t) - q_j(0) \right|,
\]
where $q_j(t)$ denotes the position of joint $j$ at time $t$. The reported curve averages this quantity across all real-world tasks. This metric indicates when measurable physical motion begins relative to the start of the instruction.

\section{Projection-Head and Consensus-Guided Execution Overhead}
\label{app:focus-head-overhead}

\paragraph{Projection-head measurement.}
The image and language projection heads contain $2.36$M parameters in total, which is small relative to the frozen $\pi_{0.5}$ backbone. In a separate profiling run, the projection heads added $0.232$\,ms per inference step on LIBERO over an average backbone forward time of $65.84$\,ms, corresponding to $0.35\%$ of the backbone forward cost. On VLA-Arena, they added $0.239$\,ms over a $65.98$\,ms backbone forward, corresponding to $0.36\%$.

In addition to the projection heads, \method{} aggregates backbone features using the shared layer weights, predicts readiness, extracts region hypotheses, and computes region-conditioned action chunks. These costs are included in Wall under the timing definition in \linkedref{Appendix}{app:eval_metrics}; the projection-head percentages above do not measure the total overhead of \method. Each action-agreement test, including repeated tests, reuses chunks from the same planning decision and adds no policy forward. After execution of a shared action prefix, further policy and region-conditioned forwards are skipped until a new word arrives.

\section{Qualitative Vision-Language Focus Maps}
\label{app:qualitative}

\begin{figure}[H]
\centering
\includegraphics[width=\linewidth]{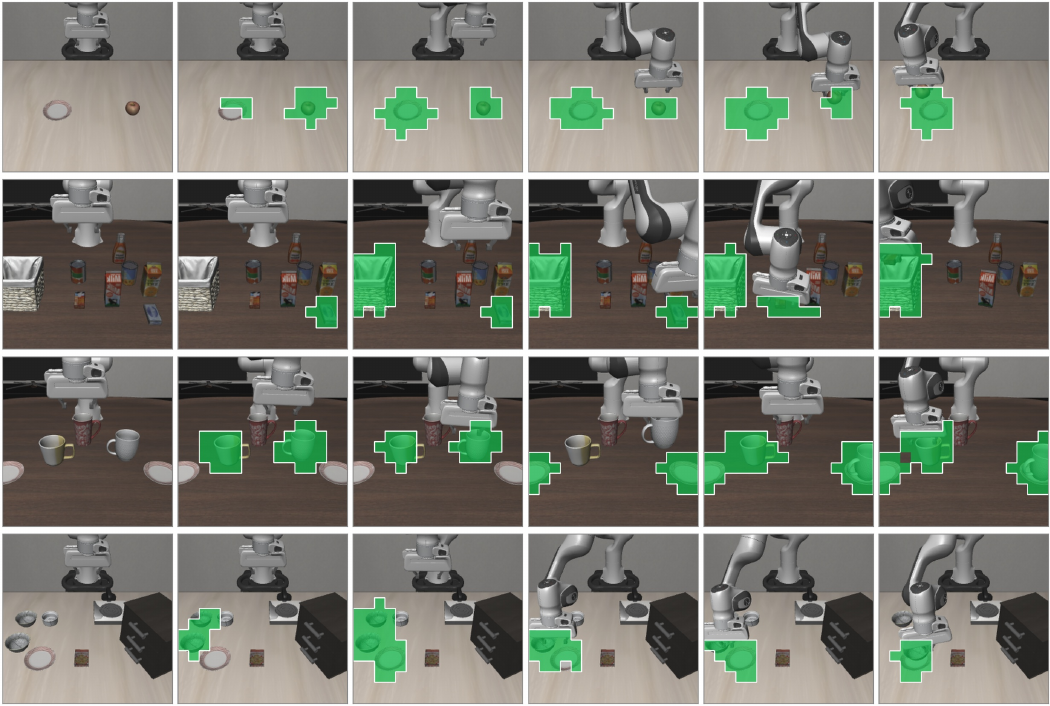}
\caption{\textbf{Qualitative vision-language focus maps in simulation.} Each row shows successive RGB observations with the predicted focus map overlaid. For visualization, only patches with focus score $p_i \geq 0.8$ are displayed; this display threshold is separate from the relative threshold used to construct region hypotheses in \linkedref{Appendix}{app:consensus_regions}.}
\label{fig:qualitative-sim}
\end{figure}

\clearpage
\vspace*{4cm}
\begin{figure}[H]
\centering
\includegraphics[width=\linewidth]{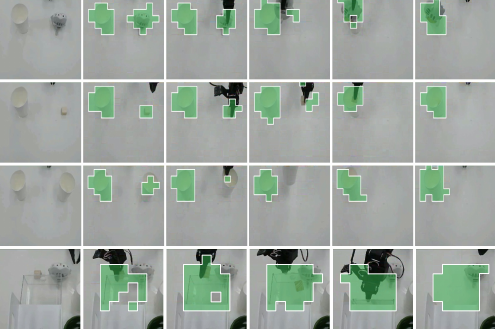}
\caption{\textbf{Qualitative vision-language focus maps on the real robot.} Each row shows successive RGB observations with the predicted focus map overlaid. For visualization, only patches with focus score $p_i \geq 0.8$ are displayed; this display threshold is separate from the relative threshold used to construct region hypotheses in \linkedref{Appendix}{app:consensus_regions}.}
\label{fig:qualitative-real}
\end{figure}

\linkedref{Figures}{fig:qualitative-sim} and~\ref{fig:qualitative-real} show focus maps after the streaming prefix has identified the target referent. Across both simulation and real-robot environments, the learned focus map often extends beyond the literal instance-mask boundary to nearby regions such as handles, lids, or contact surfaces.

\section{Qualitative Streaming Rollout Examples}
\label{app:rollout-example}

\begin{figure}[H]
\centering
\includegraphics[width=\linewidth]{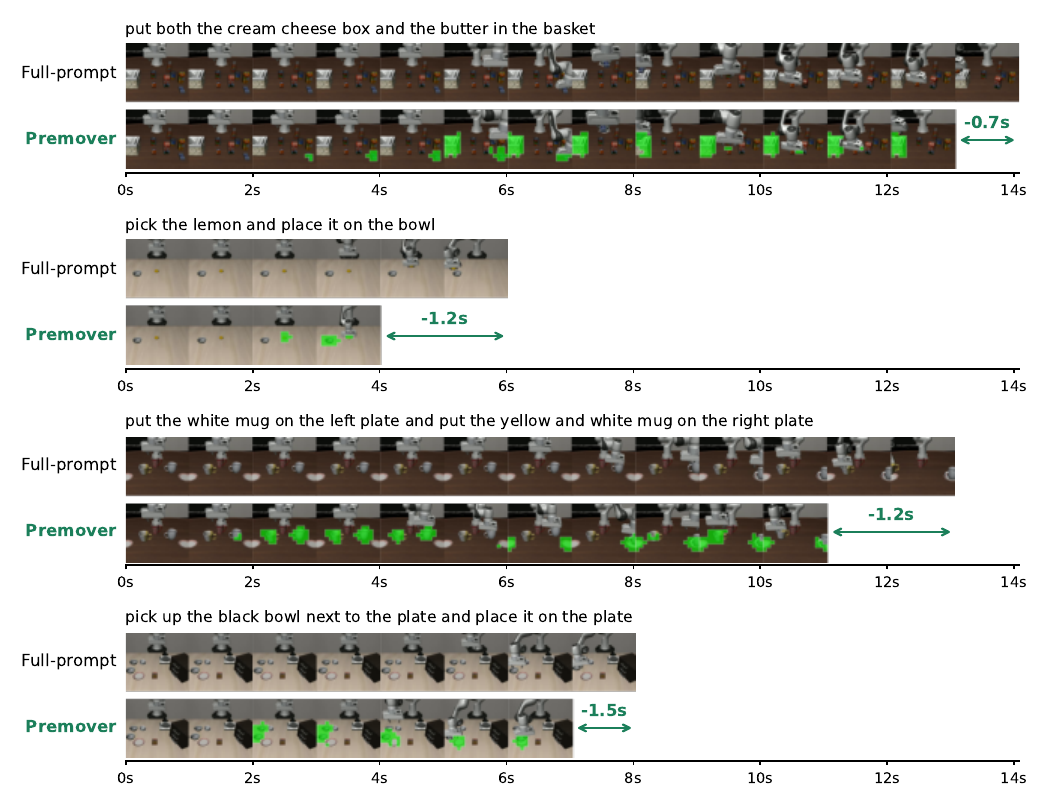}
\caption{\textbf{Qualitative streaming rollout in simulation.} Full-prompt and \method{} rollouts are shown on a common time axis aligned to the beginning of instruction delivery. Full-prompt waits for the complete instruction before execution, whereas \method{} can begin acting once the current streaming prefix contains sufficient information.}
\label{fig:rollout-sim}
\end{figure}

\clearpage
\vspace*{4cm}
\begin{figure}[H]
\centering
\includegraphics[width=\linewidth]{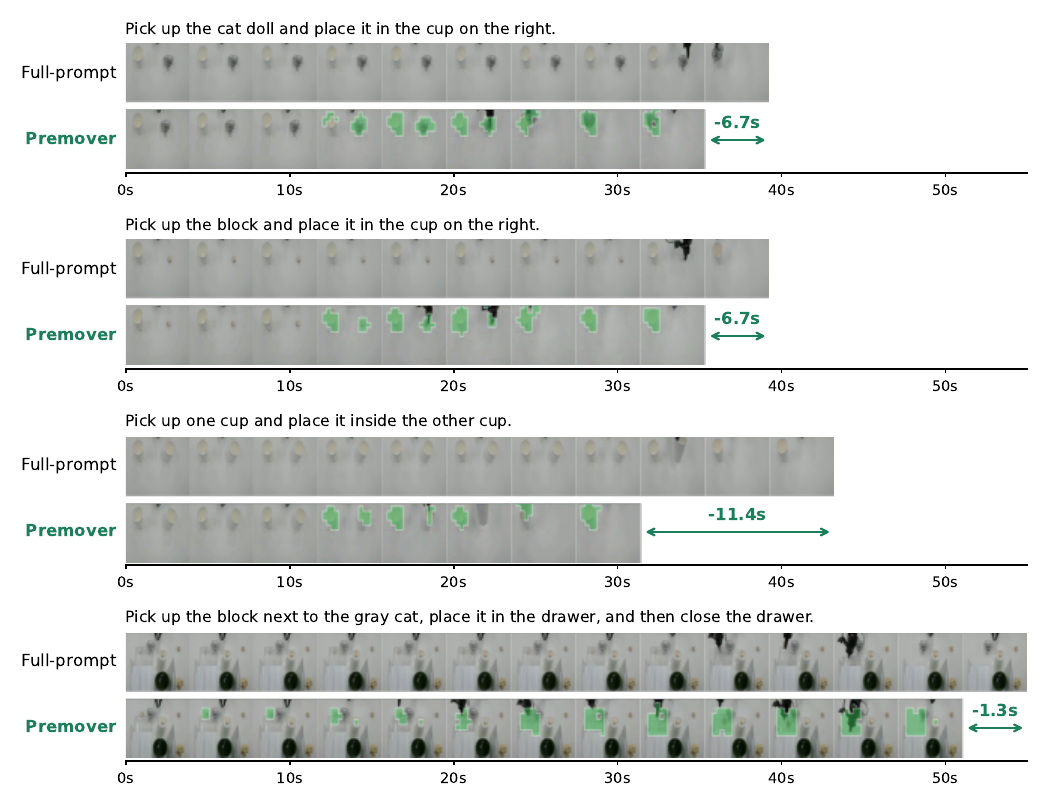}
\caption{\textbf{Qualitative streaming rollout on the real robot.} Full-prompt and \method{} rollouts are shown on a common time axis aligned to the beginning of instruction delivery. \method{} begins execution from the streaming instruction while Full-prompt waits for the complete instruction.}
\label{fig:rollout-real}
\end{figure}

\linkedref{Figures}{fig:rollout-sim} and~\ref{fig:rollout-real} illustrate how \method{} overlaps instruction delivery with robot execution in simulation and on the real robot. Because \method{} can begin execution before the complete instruction arrives, its trajectory can advance while Full-prompt is still waiting. The two methods therefore reach different robot states at the same elapsed time, and subsequent observations need not match frame by frame.

\end{document}